\documentclass[10pt,twocolumn,letterpaper]{article}

\usepackage{wacv}
\usepackage{times}
\usepackage{epsfig}
\usepackage{graphicx}
\usepackage{amsmath}
\usepackage{amssymb}

\usepackage{color}

\newcommand{\revise}[1]{{\color{black}#1}}

\usepackage{subcaption}
\usepackage{floatrow}
\usepackage{stmaryrd}
\usepackage{placeins}
\usepackage{flushend}   

\makeatletter
\newcommand{\printfnsymbol}[1]{%
  \textsuperscript{\@fnsymbol{#1}}%
}
\makeatother

\def\wacvPaperID{636} 

\wacvfinalcopy 
\usepackage{floatrow}
\usepackage{stmaryrd}
\usepackage{placeins}
\ifwacvfinal
\def\assignedStartPage{9876} 
\fi

\ifwacvfinal
\usepackage[breaklinks=true,bookmarks=false]{hyperref}
\else
\usepackage[pagebackref=true,breaklinks=true,colorlinks,bookmarks=false]{hyperref}
\fi

\ifwacvfinal
\else
\fi

\begin{document}

\title{GraphTCN: Spatio-Temporal Interaction Modeling \\for Human Trajectory Prediction}

\author{Chengxin Wang\thanks{These authors contributed equally to this work.} \qquad Shaofeng Cai\printfnsymbol{1} \qquad Gary Tan \\
National University of Singapore\\
{\tt\small \{wangcx, shaofeng, gtan\}@comp.nus.edu.sg}
}


\maketitle

\begin{abstract}

Predicting the future paths of an agent's neighbors accurately and in a timely manner is central to the autonomous applications for collision avoidance.
Conventional approaches, e.g., LSTM-based models, take considerable computational costs in the prediction, especially for the long sequence prediction.
To support more efficient and accurate trajectory predictions, we propose a novel 
CNN-based spatial-temporal graph framework GraphTCN, which models the spatial interactions as social graphs and captures the spatio-temporal interactions with a modified temporal convolutional network.
\revise{In contrast to conventional models, both the spatial and temporal modeling of our model are computed within each local time window.}
Therefore, it can be executed in parallel for much higher efficiency, and meanwhile with accuracy comparable to best-performing approaches.
Experimental results confirm that our model achieves better performance in terms of both efficiency and accuracy as compared with state-of-the-art models on various trajectory prediction benchmark datasets.


\end{abstract}
\section{Introduction}
\label{sec:introduction}

Trajectory prediction is a fundamental and challenging task, which needs to forecast the future path of the agents in autonomous applications, such as autonomous vehicles, socially compliant robots, agents in simulators, to navigate in a shared environment. 
With multi-agent interaction in these applications, the agents are required to respond timely and precisely to the environment for collision avoidance.
Therefore, the ability of the agents to predict the future paths of their neighbors in an efficient and accurate manner is thus much needed.
Although recent works~\cite{liang2019peeking,sadeghian2019sophie,huang2019stgat,mohamed2020social} have achieved great improvement in modeling complex social interactions among agents to generate accurate future paths, trajectory prediction is still a challenging task, where the deployment of the prediction models in real-world applications is mostly restricted by its high computational cost and long inference time.
For example, some small robots are only equipped with limited computing devices that can not afford the high inference cost with existing solutions.

\begin{figure}[t!]
\begin{center}
  \includegraphics[width=0.65\linewidth]{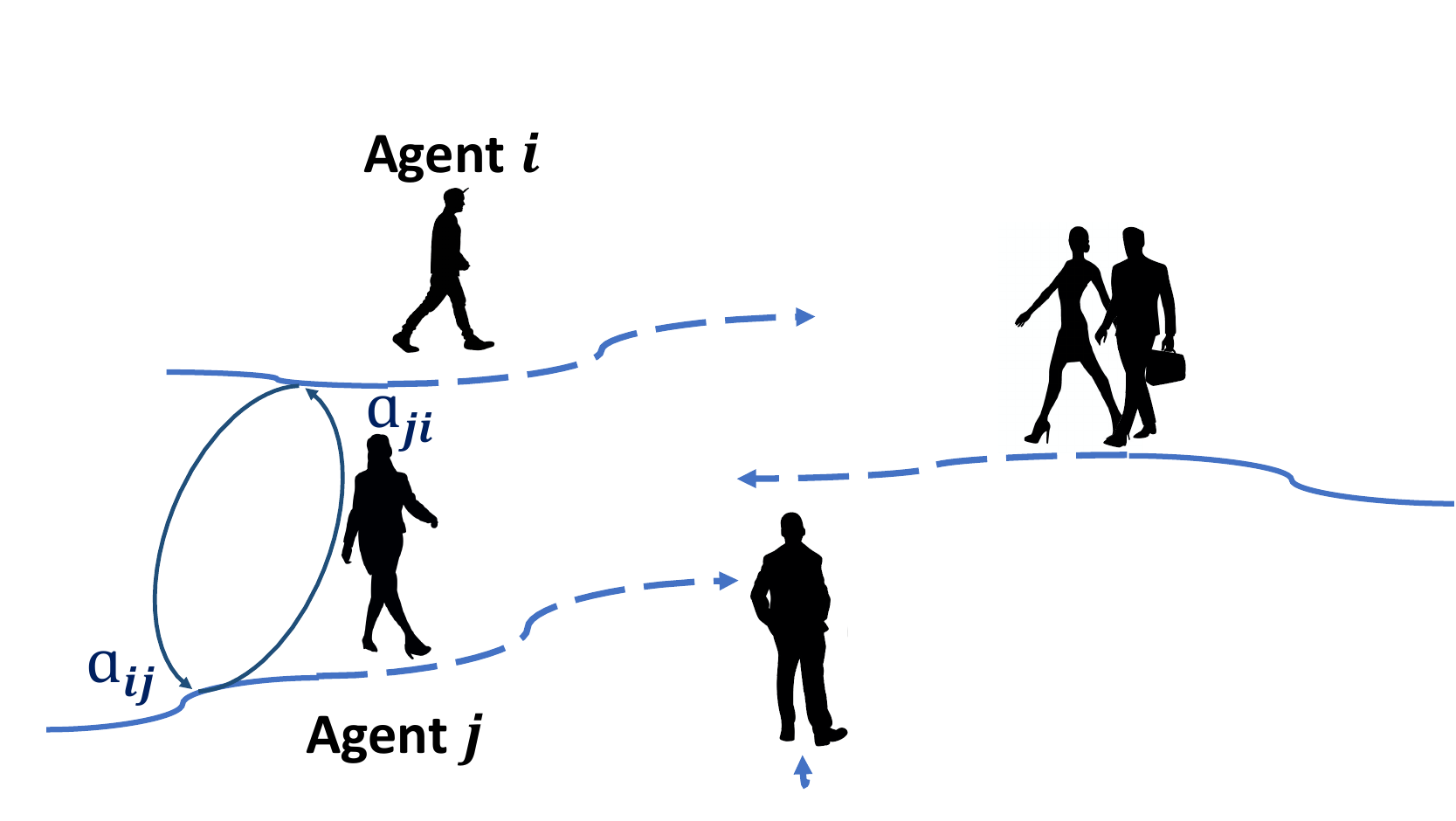}
\end{center}
\caption{
\revise{The illustration of trajectory prediction in a crowd.}
The solid blue lines are the observed trajectories, and the dash blue lines indicate the plausible future path.
The influence levels $\alpha$ between two agents are different based on their relative movement trends, e.g., $\alpha_{ij}$ and $\alpha_{ji}$ between agent $i$, and agent $j$ are different.
}
\label{fig:crowd_illustration}
\end{figure}

In particular, the trajectory prediction is typically modeled in two dimensions, i.e., the temporal dimension and the spatial dimension, which is illustrated in Fig.~\ref{fig:crowd_illustration}.
The temporal dimension models the historical movement dynamics for each agent.
Most of the state-of-the-art approaches~\cite{alahi2016social,gupta2018social,liang2019peeking,huang2019stgat,ivanovic2019trajectron,salzmann2020trajectron++} have focused on Recurrent Neural Networks (RNNs), 
to capture such sequence dynamics since RNNs are designed for sequence modeling.
However, besides the training difficulties of gradient vanishing and exploding~\cite{razvan13rnn} in modeling long sequential data, both training and inference of RNN models are notoriously slow compared with their feed-forward counterparts, e.g., Convolutional Neural Networks (CNNs).
This is largely due to the fact that each hidden state of RNNs is dependent on the previous inputs and hidden states.
As a consequence, the prediction of RNNs is produced sequentially, and thus not parallelizable.

The spatial dimension models the human-human interaction, i.e., interactions between the agent and its neighbors.
There are mainly three categories of methods proposed to capture the spatial interaction, including the distance-based~\cite{alahi2016social,gupta2018social,liang2019peeking}, 
attention-based~\cite{sadeghian2019sophie,fernando2018soft,vemula2018social,ivanovic2019trajectron} and graph-based~\cite{huang2019stgat,kosaraju2019social,zhang2019sr,mohamed2020social} approaches.
Distance-based approaches introduce a social pooling layer to summarize the crowd interactions, while the attention-based approaches instead dynamically generate the importance of neighbors using soft attention.
The graph-based approaches model the agents' representation with a graph and utilize graph neural network, e.g., GCN~\cite{kipf2017semi,zhang2019sr,mohamed2020social} or GAT~\cite{velivckovic2017graph}, to capture the spatial interaction features of agents, which is empirically more intuitive and effective in modeling complex social interactions.
However, existing graph-based approaches are mostly based on the simple aggregation of neighbor features or their absolute geometric distance, which neglects the relative relation between agents in the spatial modeling.

To improve effectiveness and efficiency, we propose a novel \textbf{\underline{Graph}}-based \textbf{\underline{T}}emporal \textbf{\underline{C}}onvolutional \textbf{\underline{N}}etwork
(GraphTCN), to capture the spatial and temporal interaction for trajectory prediction.
In the temporal dimension, different from RNN-based methods, we adopt a modified gated convolutional network (TCN) to capture the temporal dynamics for each agent.
The gated highway mechanism introduced to CNNs dynamically regulate the information flow by focusing on more salient features, and the feed-forward nature of CNN makes it more tractable in training and parallelizable for much higher efficiency both in training and inference.
In the spatial dimension, we propose an edge feature based graph attention network (EFGAT) with skip connections and the gate mechanism for each time step to model the spatial interaction between the agents.
Specifically, nodes in the graph represent agents, and edges between agents denote their relative spatial relation.
EFGAT learns the adjacency matrix, i.e., the spatial interaction, of the graph adaptively.
Together, the spatial and temporal modules of GraphTCN support more effective and efficient modeling of the interactions within each time step between agents and across the time steps for each and every agent.
We summarize our main contributions as follows:

\begin{itemize}
  \item We propose an edge feature based graph attention network (EFGAT), which introduces the relative spatial relation as prior knowledge, to capture the spatial interaction adaptively with attention.
  
  \item We propose to model the temporal interactions with a gated convolutional network (TCN), which empirically proves to be more efficient and effective.
  
  \item Our spatial-temporal framework achieves better performance compared with best-performing approaches. Specifically, we reduce the average displacement error by 19.4\% and final displacement error by 13.6\% with 5 times less predicted paths, and achieves up to 5.22x wall-clock time speedup over existing solutions.
  
\end{itemize}
We organize this paper as follows: in Section~\ref{sec:relaed_work}, we introduce the background and discuss related works in detail. Our GraphTCN framework is introduced in Section~\ref{sec:methodology}. Then in Section~\ref{sec:experiment}, results of GraphTCN measured in both accuracy and efficiency are compared with state-of-the-art approaches. Finally, Section~\ref{sec:conclusion} concludes the paper.

\section{Related Work}
\label{sec:relaed_work}

\noindent
\textbf{Human-Human Interactions.} 
Research in the crowd interaction model can be traced back to the Social Force model~\cite{helbing1995social}, which adopts the nonlinearly coupled Langevin equations to represent the attractive and repulsive forces for human movement in the crowed scenarios. 
Similar hand-crafted approaches\cite{treuille2006continuum,antonini2006discrete,wang2007gaussian} have proved successful in crowd simulation~\cite{hou2014social,saboia2012crowd}, crowd behavior detection~\cite{mehran2009abnormal}, and trajectory prediction~\cite{yamaguchi2011you}.
Recent works instead investigate deep learning techniques to capture the interaction between agents and neighbors.
The distance-based approaches~\cite{alahi2016social,gupta2018social,mangalam2020not} either adopt the grid-based pooling or symmetric function to aggregate the hidden states from neighbors or encode the geometric relation between the agents.
Different from the distance-based methods, attention-based approaches~\cite{sadeghian2019sophie,vemula2018social,fernando2018soft,zhang2019sr} provide better crowd modeling since they differentiate the importance of neighbors by soft attention or gating mechanisms.
More recent works~\cite{huang2019stgat,kosaraju2019social,mohamed2020social} adopts graph-based networks to learn the social interaction by aggregating neighborhood features adaptively with the adjacency matrix, which provides an effective way to represent the pedestrian's topology in a shared space.
Social-STGCNN~\cite{mohamed2020social} captures the spatial relation by introducing a kernel function on the weighted adjacency matrix;
STGAT~\cite{huang2019stgat,kosaraju2019social} adopts GAT directly on the LSTM hidden states to capture the spatial interaction between pedestrians.
However, Social-STGCNN only focuses on distance features between agents, and STGAT simple aggregates neighbor features. 
EGNN~\cite{gong2019exploiting} incorporates the edge feature into the graph attention mechanism to exploit richer graph information.
However, EGNN neglects the relative relation between pedestrians.
We propose to model the pedestrian interactions with a novel edge feature based graph network, which integrates the relative distance feature into graph attention to learn an adaptive adjacency matrix for the most salient interaction information.

\noindent
\textbf{Pattern-based Sequence Prediction.}
Sequence prediction refers to the problem of predicting the future sequence using historical information.
Recently, pattern-based methods prevails for many sequence prediction tasks, e.g., speed recognition~\cite{oord2016wavenet,chorowski2014end,graves2014towards}, activity recognition~\cite{donahue2015long,ibrahim2016hierarchical}, and natural language processing~\cite{cho2014learning,sutskever2014sequence,gehring2017convolutional}.
In particular, trajectory prediction can be formulated as a sequence prediction task, which uses historical movement patterns of the agent to predict the future path.
Most trajectory prediction methods adopt recurrent neural networks (RNNs), e.g., Long Short-Term Memory (LSTM) networks~\cite{hochreiter1997long}, to capture the temporal movement.
However, RNN-based models suffer from gradient vanishing and exploding in training and focus more on recent inputs during prediction, especially for long input sequences.
Many sequence prediction works~\cite{oord2016wavenet,wu2019graph} instead adopt convolutional neural networks (CNNs).
The convolutional networks can effectively capture long-term dependency and greatly improve prediction efficiency.
The superiority of CNN-based methods can be largely attributed to the convolutional operation, which is independent of preceding time-steps and thus can process in parallel.
The recent work~\cite{nikhil2018convolutional} proposes a compact CNN model to capture the temporal information, and the results confirm that the CNN-based model can yield competitive performance in trajectory prediction.
However, it fails to model the spatial interaction between pedestrians.
In this work, we propose to capture the spatial interaction with EFGAT and introduce gated convolutional networks to better capture the temporal dynamics.


\noindent
\textbf{Graph Networks for Trajectory Prediction.}
Many studies adopt spatial-temporal graph neural networks (STGNNs) for the sequence prediction task, such as action recognition~\cite{yan2018spatial,si2019attention}, 
taxi demand prediction~\cite{yao2018deep}, and traffic prediction~\cite{yao2018modeling}.
Specifically, the sequence can be formulated as a sequence of graphs of nodes and edges, where nodes correspond to agents and edges denote their interactions.
The sequence can then be effectively modeled with the spatial-temporal graph network.
Likewise, the trajectory prediction task can be modeled with the \revise{spatial-temporal graph network~\cite{vemula2018social,wang2019pedestrian,huang2019stgat,liang2020garden,mohamed2020social}}.
In particular, the prediction task needs to be modeled in two dimensions, i.e., the spatial dimension and the temporal dimension.
The spatial dimension models the interaction between the agent and its neighbors, and the temporal dimension models the historical trajectory for each agent.
Specifically, each node in the graph represents one pedestrian of a scene, and each edge between two nodes captures the interaction between the two corresponding pedestrians.
For example, social attention~\cite{vemula2018social} models each node with the location of the agent, and edge with the distance between pedestrians, where the spatial relation is modeled with an attention module and then the temporal with RNNs.
Similarly, ~\cite{wang2019pedestrian} constructs the STGNN with Edge RNN and Node RNN based on the location.
STGAT~\cite{huang2019stgat} adopts GAT to capture the spatial interaction by assigning different importance to neighbors and adopts extra LSTMs to capture the temporal information.
The major limitation of these methods lies in capturing the spatial interaction along the temporal dimension.
Notably, the future path of an agent is not only dependent on the current position but meanwhile the neighbors'.
However, the information of such spatial interaction may be lost during the aggregation of the node features along the temporal dimension using RNN-based models.
Different from RNN-based methods, Social-STGCNN~\cite{mohamed2020social} and
Graph WaveNet~\cite{wu2019graph} adopts CNNs to alleviate parameter inefficiency and demonstrate the capability of CNNs in the temporal modeling of long sequences.
In this paper, we propose an enhanced temporal convolutional network to integrate both the temporal dynamics of the agent and the spatial interactions.

\section{GraphTCN}
\label{sec:methodology}

The goal of trajectory prediction is to predict the future paths of all agents that are present in a scene.
Naturally, the future path of an agent depends on its historical trajectory, i.e., the temporal interaction, and is influenced by the trajectories of neighboring agents, i.e., the spatial interaction.
Therefore, the trajectory prediction model needs to take into account both features when modeling the spatial and temporal interactions for the prediction.

\noindent
\textbf{Problem Formulation}
Formally, the trajectory prediction can be defined as follows:
given N pedestrians observed in a scene with $T_{obs}$ steps,
the position of a single pedestrian $i \in \{1, \dots, N\}$ at the time step $t \in \{1, \dots, T_{obs}\}$ is denoted as $\mathbf{X}_{i}^{t} = 
\left(x_{i}^{t}, y_{i}^{t}\right)$.
Therefore, the observation positions of the pedestrian $\mathbf{X}_i$ can be represented as $\mathbf{X}_{i}^{1:T_{obs}} = [\mathbf{X}_{i}^{1}$; $\mathbf{X}_{i}^{2}$;...; $\mathbf{X}_{i}^{T_{obs}}]$.
The goal of trajectory prediction is to predict all the future positions $\hat{\mathbf{Y}}_{i}^{t}$ ($t \in \{T_{obs+1}, \dots, T_{obs+pred}$\}) concurrently.

\subsection{Overall Framework}

\begin{figure*}
\begin{center}
\begin{subfigure}{0.76\textwidth}
  \centering
    \includegraphics[height=5.6cm]{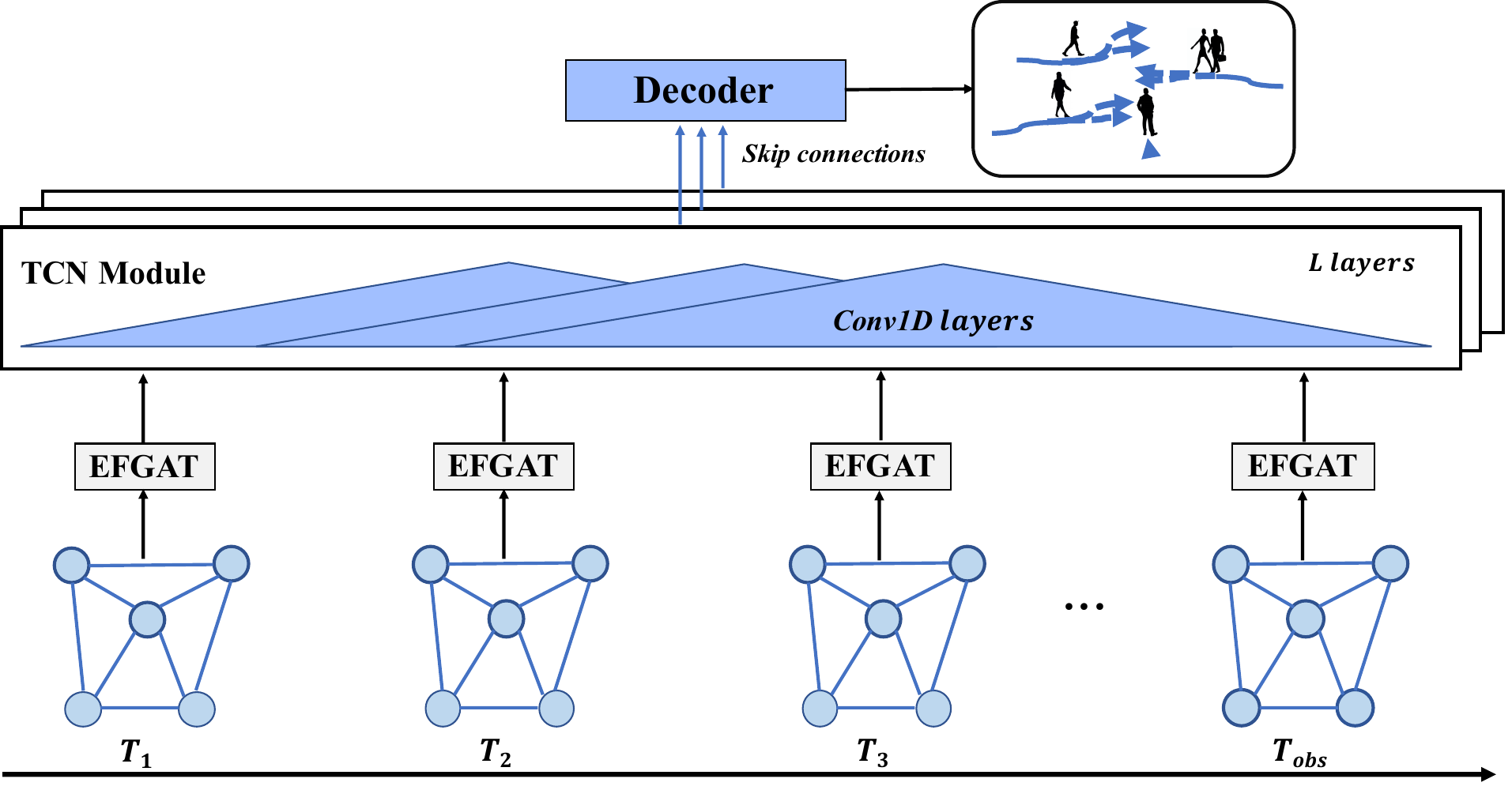}
  \caption{GraphTCN Overview}
  \label{fig:overview}
\end{subfigure}
\begin{subfigure}{.23\textwidth}
  \centering
  \includegraphics[height=5cm]{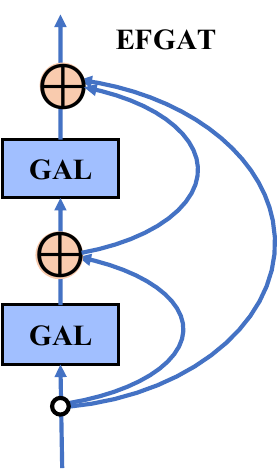} 
  \caption{\revise{EFGAT}}
  \label{fig:efgat}
\end{subfigure}
\end{center}

\caption{(a) The overview of GraphTCN, where EFGAT captures the spatial interaction between agents for each time step and is based on the historical trajectory embedding. TCN further captures the temporal interaction across time steps.
The decoder module then produces multiple socially acceptable trajectories for all the agents simultaneously.
(b) EFGAT captures the spatial salient information with graph attentional layers (GAL) and skip connections.
}
\label{fig:overall_framework}
\end{figure*}

As illustrated in Fig.~\ref{fig:overall_framework}(a), GraphTCN comprises three key modules, including the edge feature graph attention (EFGAT) module, temporal convolutional (TCN) module, and a decoder.
First, we embed the absolute positions and relative positions of each pedestrian into a fixed-length hidden space and feed these trajectories features into the EFGAT module.
The residual learning mechanism and skip connections~\cite{he2016deep} are incorporated into the network to facilitate the gradient backpropagation and encourage the intermediate feature reusage.
The TCN module is a feed-forward one-dimensional convolutional network with a gating activation unit~\cite{oord2016wavenet} for capturing the most salient features.
Finally, the decoder module produces future trajectories of all pedestrians.
We elaborate on the details of each module of GraphTCN in the following sections.

\subsection{EFGAT Module for Spatial Interaction}

The EFGAT module shown in Fig.~\ref{fig:overall_framework}(b) is designed to encode the spatial interaction between pedestrians with graph attentional layers and graph residual connections.
Formally, pedestrians within the same time step can be formulated as a directed graph $\mathcal{G}=(\mathcal{V}^t, \mathcal{E}^t)$, where each node $v_{i}^t \in \mathcal{V}^t, i \in \{1, \dots, N\}$ corresponds to the $i$-th pedestrian, and the weighted edge $\left(v_{i}^t, v_{j}^t\right) \in \mathcal{E}^t$ represents the human-human interaction from pedestrian $i$ to $j$.
The adjacency matrix $\mathbf{A}^t \in \mathbb{R}^{N \times N}$ of $\mathcal{G}$ thus represents the spatial relationships between pedestrians.

\begin{figure}[!htb]
\begin{center}
  \includegraphics[width=0.55\linewidth]{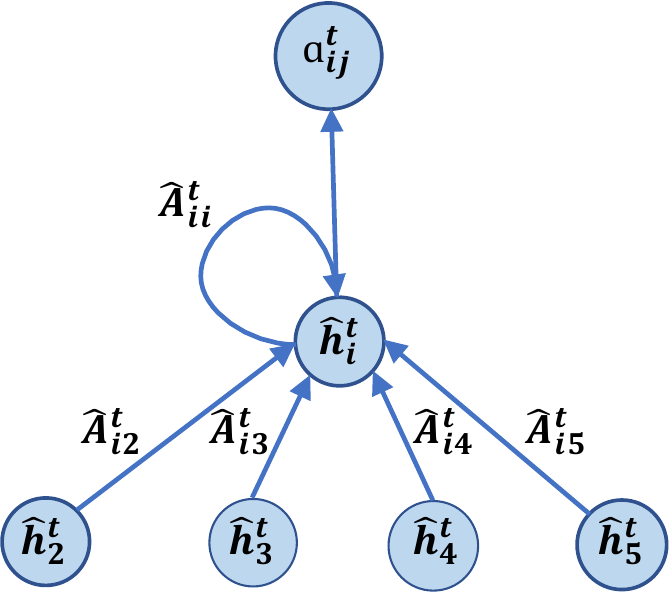}
\end{center}
  \caption{An illustration of the graph attentional layer with 5 nodes employed by our EFGAT module. The attention $\alpha_{i j}^{t}$ between node $i$ and its neighbors is learned from their embedding features and relative spatial relation $\hat{\mathbf{A}}_{ij}^{t}$. 
  }
\label{fig:single_layer_egat}
\end{figure}

We represent the spatial relation of nodes as an asymmetric, non-negative matrix in this task since the influence between the agents should be different based on their relative movement behavior.
Therefore, instead of constructing graphs with undirected spatial distance, we introduce relative spatial location as prior edge feature knowledge of the adjacency matrix:

\begin{equation}
\hat{\mathbf{A}}_{ij}^{t}=\phi_{s}\left(x_{i}^{t}-x_{j}^{t}, y_{i}^{t}-y_{j}^{t}; \mathbf{W}_{s}\right)
\end{equation}

\noindent
\revise{
where $\phi_s(\cdot)$ embeds the relative distance features to a higher dimension $\mathbb{R}^{F_1}$ with a linear transformation,} and $\mathbf{W}_{s}$ is the embedding weight.
We feed the learnable edge weights and node features into graph attentional layers illustrated in Fig.~\ref{fig:single_layer_egat} to capture the spatial interaction:

\begin{equation}
\alpha_{i j}^{t}=\frac{\exp \left( \sigma \left( \mathbf{h}^T_{\alpha} \left(\mathbf{W}_{h} \hat{\mathbf{h}}_{i}^{t}+\mathbf{W}_{h} \hat{\mathbf{h}}_{j}^{t}+\hat{\mathbf{A}}_{i j}^{t}\right)\right)\right)}{\sum_{k \in \mathcal{N}_{i}} \exp \left( \sigma \left( \mathbf{h}^T_{\alpha} \left(\mathbf{W}_{h} \hat{\mathbf{h}}_{i}^{t}+\mathbf{W}_{h} \hat{\mathbf{h}}_{k}^{t}+\hat{\mathbf{A}}_{i k}^{t}\right)\right)\right)}
\end{equation}


\noindent
where $\hat{\mathbf{h}}_i^t \in \mathbb{R}^{F_0}$ is the node input feature for pedestrian $i$ at time step $t$, ${F_0}$ is the dimension of node features, \revise{${\cal N}_i$ is the set of neighbors for node $i$ in the graph}, $\sigma(\cdot)$ is the LeakyReLU activation, 
and $\mathbf{W}_{h} \in \mathbb{R}^{F_1 \times F_0}$
and $\mathbf{h}_{\alpha} \in \mathbb{R}^{F_1}$ are learnable weights.
In this way, $\alpha_{i j}^{t}$ determines the importance weight of neighbor $j$ to pedestrian $i$ dynamically via the self-attention mechanism.
The gating function empirically proves to be effective in controlling the signal bypassing~\revise{~\cite{oord2016wavenet,xie2018rethinking,dauphin2017language}}.
We therefore adopt the gated activation unit to dynamically regulate the information flow and select salient features:

\begin{equation}
\mathbf{g}_{i}^{t}= g\left(\mathbf{W}_{h} \hat{\mathbf{h}}_{i}^{t}+b_{h}\right) \odot \left(\mathbf{W}_{h} \hat{\mathbf{h}}_{i}^{t}+b_{h}\right)
\end{equation}

\noindent
where $g(\cdot)$ is the \textit{tanh} activation function,
$b_{h}$ is the bias, and $\odot$ denotes the element-wise multiplication.
This can be understood as a multiplicative skip connection which facilitates gradients
flow through layers~\cite{dauphin2017language}.
To stabilize the self-attention process~\cite{velivckovic2017graph,wu2019graph}, we adopt the multi-head attention mechanism:

\begin{equation}
\revise{
{h}_{i}^{t}= \bigparallel_{k=1}^{K} \sigma\left(\sum_{j \in \mathcal{N}_{i}} \alpha_{k i j}^{t} \mathbf{g}_{j}^{t}\right) + \mathbf{R}({\hat{\mathbf{h}}}_{i}^{t}; \mathbf{W}_{r})
}
\end{equation}

\noindent
where 
$\mathbf{W}_{r}$ is the learnable parameters, $\bigparallel$ denotes concatenation, and $K$ is the number of attention heads.
$\mathbf{R}(\cdot)$ denotes the graph residual term~\cite{oord2016wavenet,velivckovic2017graph,wu2019graph}.
\revise{We name the proposed multi-head graph attention layers as GAL, which
can be stacked multiple times for better modeling spatial relation (e.g., twice as in Fig.~\ref{fig:overall_framework}(b)).
Subsequently, we can obtain final node representations of ${\mathbf{h}}=\left\{{\mathbf{h}}_{1}, {\mathbf{h}}_{2}, \ldots, {\mathbf{h}}_{N} \right\}$, where ${\mathbf{h}}_{i} \in \mathbb{R}^{T_{obs} \times (K \cdot F_1)}$ captures the aggregated spatial interaction between pedestrian $i$ and all the neighbors at each time step.}
Therefore, the EFGAT module can learn a self-adaptive adjacency matrix that captures the relative importance for different pedestrians.

\subsection{TCN for Spatial and Temporal Interaction Modeling}

\begin{figure}[!htb]
\begin{subfigure}{.65\textwidth}
\centering
  \includegraphics[height=3.6cm]{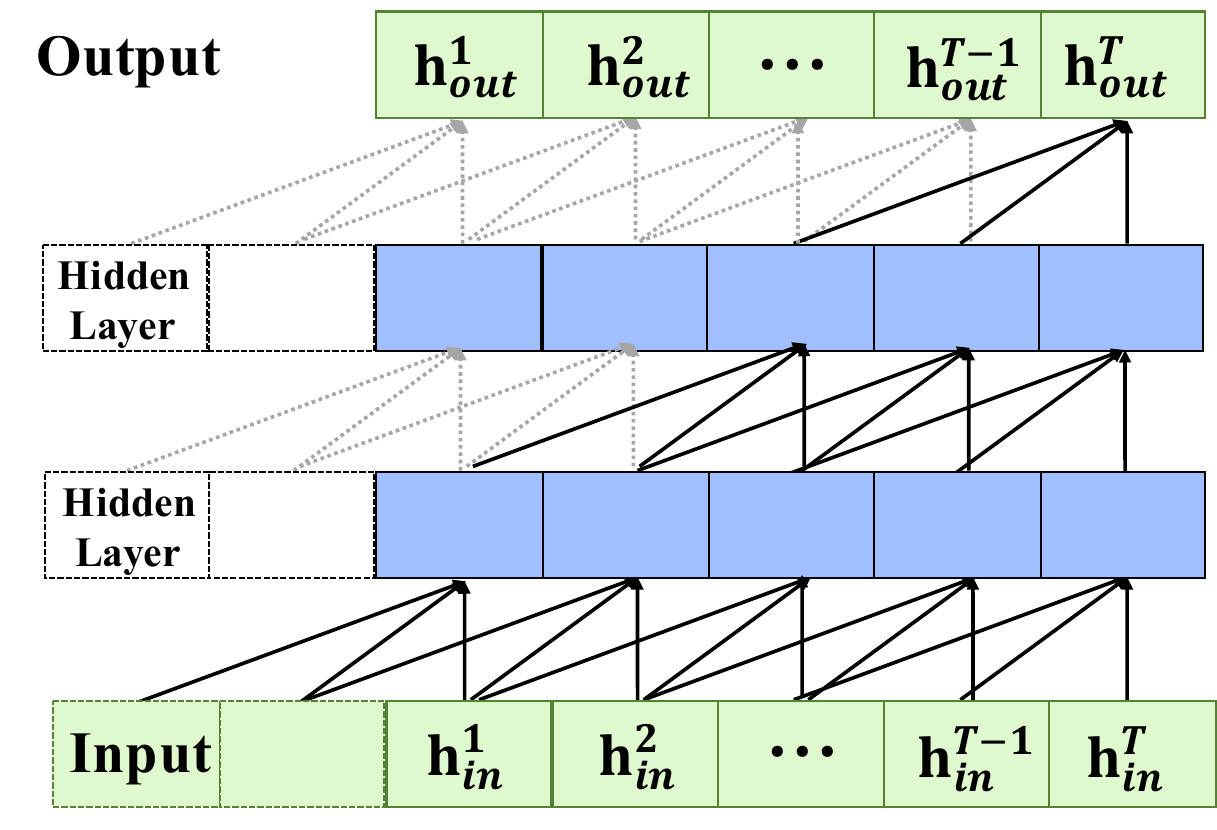} 
  \caption{}
  \label{fig:tcn_layers}
\end{subfigure}
\begin{subfigure}{.34\textwidth}
  \centering
  \includegraphics[height=3.6cm]{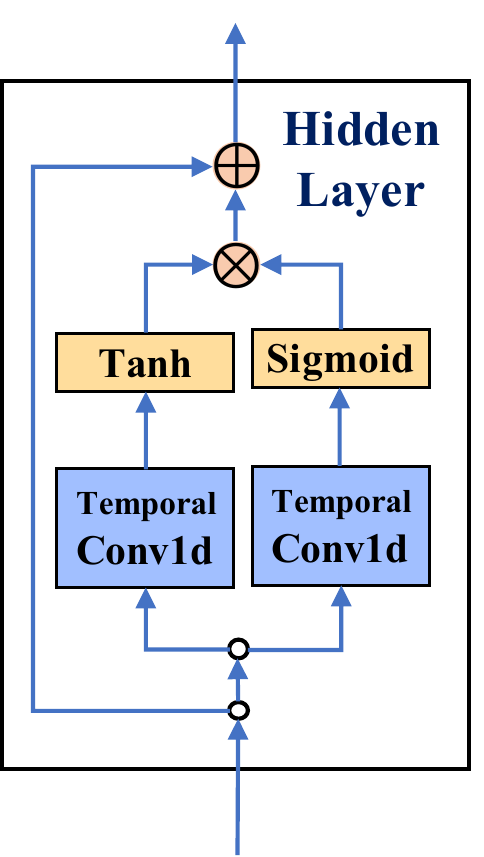} 
  \caption{}
  \label{fig:tcn_layer}
\end{subfigure}
 \caption{
(a) An illustration of TCN with a stack of 3 convolution layers of kernel size 3.
The input $\mathbf{h}_{in}$ (i.e., $\mathbf{h}^{(0)}$) contains the spatial information captured by the preceding EFGAT modules.
The output of TCN $\vec{\mathbf{h}}$ is collected by concatenating $\mathbf{h}_{out}$ (i.e., $\mathbf{h}^{(L)}$) across time.
(b) The gating function in each of the TCN layers to control the bypass signals.
}
\label{fig:convolutional_layers}
\end{figure}

The movement pattern of a pedestrian is greatly influenced by the historical trajectory and the moving patterns of neighboring pedestrians.
We therefore propose to capture the spatial and temporal interaction between pedestrians using a modified temporal convolution network (TCN), which is illustrated in Fig.~\ref{fig:convolutional_layers}.

The network shown in Fig.~\ref{fig:convolutional_layers}(a) can be regarded as a short-term and long-term encoder, where lower convolution layers focus on local short-term interactions, while in higher layers, long-term interactions are captured with a larger receptive field.
For example, if the kernel size of the TCN is $k$, the receptive field size in the $l$-th layer is $(k-1) \cdot l+1$, which increases linearly ascending layers.
Therefore, the top layer of TCN captures interactions within a longer time span.
Since the order of the input is important in the sequence prediction task, we therefore adopt the left padding of size $k-1$ instead of symmetric padding for the convolution, where each convolution output convolves over the input of the corresponding time step and the preceding $k-1$ time steps.
The output size of each convolution then remains the same as the input.
In each layer of TCN~\ref{fig:convolutional_layers} in Fig.(b), the gated activation unit utilizes two non-linear functions to dynamically regulate the information flow formed as:

\begin{equation}
\mathbf{h}^{(l+1)}= g\left(\mathbf{W}_{g}^{(l)} * \mathbf{h}^{(l)} \right) \odot \sigma\left(\mathbf{W}_{f}^{(l)} * \mathbf{{h}}^{(l)}\right)
\end{equation}

\noindent
where $\mathbf{h}^{(0)}$ is the output $\mathbf{h}$ from the EFGAT modules, $\mathbf{h}^{(l)} \in \mathbb{R}^{N \times T_{obs} \times F_2}$, $\mathbf{W}_{g}$ and $\mathbf{W}_{f}$ are learnable 1D-convolution parameters, and $\sigma(\cdot)$ here denotes the \textit{sigmoid} function.
\revise{The final output of the TCN module can be denoted as $\vec{\mathbf{h}} \in \mathbb{R}^{N \times T_{obs} \times F_2}$.}
In this way, the embedding vector $\vec{\mathbf{h}}_i$ captures all the spatial-temporal interaction between the $i$-th pedestrian and its neighbors.
We note that TCN can handle much longer input sequences with the dilated convolution~\cite{oord2016wavenet}, which is more efficient than RNN-based methods.

\subsection{Future Trajectory Prediction}

In real-world applications, given the historical trajectory, there are multiple plausible paths of future movements.
We also model such uncertainty of the final movement in our decoder module for the trajectory prediction.

Following STGAT~\cite{huang2019stgat}, the decoder module produces multiple socially acceptable trajectories by introducing \revise{a shared random noise $\boldsymbol{z} \in \mathbb{R}^{T_{obs} \times F_3}$, which is concatenated with the spatial-temporal embedding $\vec{\mathbf{h}}$, as part of the decoder input.}
\revise{Specifically, the input of the decoder can be denoted as $\tilde{\mathbf{h}} \in \mathbb{R}^{N \times T_{obs} \times (F_2+F_3)}$.
We adopt a canonical linear layer to generate the relative future locations $\Delta \hat{\mathbf{Y}} \in \mathbb{R}^{N \times T_{pred} \times 2}$ and denote the architecture with such a decoder as GraphTCN.
}

\revise{The predicted relative location $\Delta \hat{\mathbf{Y}}$ is the relative position to the origin for all the pedestrians.} 
We then convert relative positions to absolute positions $\hat{\mathbf{Y}}$ and adopt the variety loss as the loss function for training, which computes the minimum ADE loss among the $M$ plausible trajectories:

\begin{equation}
\mathcal{L}_{\text {ADE}} (\hat{\mathbf{Y}}) =\frac{\sum_{i=1}^{N} \sum_{t=1}^{T_{pred}}\left\|\hat{\mathbf{Y}}_{i}^{t}-\mathbf{Y}_{i}^{t}\right\|_{2}}{N T_{pred}}
\label{eq:ade}
\end{equation}

\begin{equation}
\mathcal{L}_{\text {variety}} = \min _{m}\left(\mathcal{L}_{A D E}\left(\hat{\mathbf{Y}}^{(m)}\right)\right)
\end{equation}

\noindent
where ${\mathbf{Y}}$ is the ground truth, $\hat{\mathbf{Y}}^{(1)}, \dots, \hat{\mathbf{Y}}^{(M)}$ are the $M$ plausible trajectories predicted.
Although this loss function may lead to a diluted probability density function~\cite{thiede2019analyzing}, we empirically find that it facilitates better predictions of multiple future trajectories.

We further integrate the deep generative strategy\revise{~\cite{sohn2015learning,ivanovic2019trajectron,mangalam2020not}} adopted widely in multimodal prediction to enhance the decoder of our GraphTCN.
\revise{
Specifically, during training, we concatenate $\overrightarrow{\mathbf{h}}$ with the ground-truth future trajectories encoded by an MLP layer, and then further encode the two features with an MLP to produce $\boldsymbol{\mu}$ and $\boldsymbol{\sigma}$ for the noise distribution $\boldsymbol{\hat{z}} = \mathcal{N}(\boldsymbol{\mu}, \boldsymbol{\sigma}), \boldsymbol{\hat{z}} \in \mathbb{R}^{N \times F_4}$ following CVAE~\cite{sohn2015learning,mangalam2020not}.
Note that $\boldsymbol{\hat{z}}$ is randomly sampled from distribution $\mathcal{N}(\boldsymbol{0}, \mathbf{I})$ during inference.
For the final relative location prediction, we again concatenate $\boldsymbol{\hat{z}}$ with $\vec{\mathbf{h}}$ and feed them into a linear layer to produce $\Delta \hat{Y}$.
We further introduce the KL divergence regularization term~\cite{Goodfellow-et-al-2016,lee2017desire} to stabilize the training process:}

\begin{equation}
\mathcal{L} = \lambda_{1} \mathcal{L}_{\text {variety}} + \lambda_{2} {D_{K L}(\mathcal{N}(\boldsymbol{\mu}, \boldsymbol{\sigma}) \| \mathcal{N}(\boldsymbol{0}, \mathbf{I}))}
\end{equation}

\noindent
GraphTCN with such a decoder is denoted as GraphTCN-G in the following experiments.

\section{Experiments}
\label{sec:experiment}

In this section, we evaluate our GraphTCN on two world coordinates trajectory prediction datasets, i.e., ETH~\cite{pellegrini2010improving} and UCY~\cite{lerner2007crowds}, and compare the performance of GraphTCN with state-of-the-art approaches.

\subsection{Datasets and Evaluation Metrics}

ETH and UCY datasets comprise five outdoor environments that are recorded from a fixed top-view.
The ETH dataset includes ETH and Hotel, and the UCY dataset consists of UNIV, ZARA1, and ZARA2.
In these datasets, pedestrians exhibit complex behaviors, including nonlinear trajectories, moving from different directions, walking together, walking unpredictably, avoiding collisions, standing, etc.
The crowd density of a single scene in each environment varies from 0 to 51 pedestrians per frame.
All datasets are records at 25 frames per second (FPS), and the pedestrian trajectory is extracted at every 2.5 FPS.

We use two metrics to evaluate model performance:
\textit{Average Displacement Error (ADE)} defined in Equation~\ref{eq:ade}, which is the average Euclidean distance between the predicted trajectory and the ground truth over all prediction time steps, and \textit{Final Displacement Error (FDE)}, which is the Euclidean distance between the predict position and the ground truth position at the final time step $T_{pred}$. 

The model is trained with the leave-one-out policy~\cite{alahi2016social,gupta2018social,vemula2018social,huang2019stgat}.
We produced 4 samples for the next 4.8 seconds (12 timesteps) based on 3.2 seconds (8 timesteps) observations.

\subsection{Implementation Details}
We train with Adam optimizer in 50 epochs with a learning rate of 0.0001.
The node feature embedding size is set to 64. 
The EFGAT module comprises two graph attention layers with \revise{attention heads K = 2, 1, 
$F_0$ = 64, 32,
and a output dimension $F_1$ = 16, 32 for the first and second GAL respectively.}
\revise{$F_2$ is also set to 32, and the noise $\boldsymbol{z}$ has a dimension of $F_3$ = 4.}
For the GraphTCN-G decoder, the ground truth trajectory during training is encoded 
into a dimension of 64 for $F_4$.
$M$ is set to 4 and 20 for predicting 4 and 20 sample paths.
All the LeakyReLU in our model has a negative slope of 0.2.
$\lambda_{1}$ is set to 1, and $\lambda_{2}$ is set to 0.5 for the first 15 epochs and 0.2 for the rest epochs for GraphTCN-G.

\subsection{Baselines}
We compare our framework with the baselines and four state-of-the-art approaches:
\textit{LSTM} adopts the vanilla LSTM encoder-decoder model to predict the sequence of every single pedestrian.
\textit{Social LSTM}~\cite{alahi2016social} builds on top of LSTM and introduces a social pooling layer to capture the spatial interaction between pedestrians.
\textit{CNN}~\cite{nikhil2018convolutional} adopts the CNNs to predict the sequence.
\textit{SR-LSTM}~\cite{zhang2019sr} obtains the spatial influence by iteratively refining the LSTM hidden states through the gate and attention mechanism.
\textit{Social GAN}~\cite{gupta2018social} improves over Social LSTM with socially generative GAN to generate multiple plausible trajectories.
\textit{Trajectron}~\cite{ivanovic2019trajectron} utilizes LSTM to capture the spatial and temporal relations and incorporates CVAE \revise{~\cite{sohn2015learning}} to generate the distributions of future paths.
\revise{\textit{SoPhie}~\cite{sadeghian2019sophie} introduces the social and physical attention mechanisms to an LSTM based GAN model.}
\textit{Social-STGCNN}~\cite{mohamed2020social} is one of the SOTA approaches that utilizes CNNs to extract spatio-temporal features.
\textit{STGAT}~\cite{huang2019stgat} is one of the SOTA approaches which adopts vanilla GAT to model the spatial interactions and LSTMs to capture temporal interaction.

\subsection{Quantitative Results}

The results in Table \ref{quantitative_result} show that GraphTCN achieves consistently better performance compared with existing models on these benchmark datasets.
Our model generates multiple trajectories at once for future trajectories.
\revise{We can notice that GraphTCN achieves better prediction performance than other baselines with only four predictions instead of 20 as in most baselines, e.g., STGAT~\cite{huang2019stgat}, with an ADE of 0.36 and an FDE of 0.72 on average.}
These results confirm that our GraphTCN yields competitive results even with less generated paths compared with previous approaches in terms of prediction accuracy, especially on the more complex dataset UNIV, ZARA1, and ZARA2.

\begin{table*}
\begin{center}
\caption{Quantitative results of GraphTCN compared with baseline approaches.
Evaluation metrics are reported in ADE / FDE in meters (the lower numerical result is better). 
The $\ast$ mark denotes the deterministic model, and the rest of the baseline approaches are stochastic models with M = 20 prediction samples.
}
\label{quantitative_result}
\begin{tabular}{c||c|c|c|c|c||c}
\hline
Method & ETH & HOTEL & UNIV & ZARA1 & ZARA2 & AVG\\
\hline
LSTM$\ast$~\cite{alahi2016social}& 1.09 / 2.41 & 0.86 / 1.91 & 0.61 / 1.31 & 0.41 / 0.88 & 0.52 / 1.11 &
0.70 / 1.52\\
Social-LSTM$\ast$~\cite{alahi2016social} & 1.09 / 2.35 & 0.79 / 1.76 & 0.67 / 1.40 & 0.47 / 1.00 & 0.56 / 1.17 & 0.72 / 1.54\\
CNN$\ast$~\cite{nikhil2018convolutional} & 1.04 / 2.07 & 0.59 / 1.27 & 0.57 / 1.21 & 0.43 / 0.90 & 0.34 / 0.75 & 0.59 / 1.22\\
SR-LSTM$\ast$~\cite{zhang2019sr} & 0.63 / 1.25 & 0.37 / 0.74 & 0.51 / 1.10 & 0.41 / 0.90 & 0.32 / 0.70 & 0.45 / 0.94\\
\hline
Social-GAN~\cite{gupta2018social} & 0.81 / 1.52 & 0.72 / 1.61 & 0.60 / 1.26 & 0.34 / 0.69 & 0.42 / 0.84 & 0.58 / 1.18\\
Trajectron~\cite{ivanovic2019trajectron} & 0.59 / 1.14 & 0.35 / 0.66 & 0.54 / 1.13 & 0.43 / 0.83 & 0.43 / 0.85 & 0.56 / 1.14 \\
\revise{
SoPhie~\cite{sadeghian2019sophie}} & 0.70 / 1.43 & 0.76 / 1.67 & 0.54 / 1.24 & 0.30 / 0.63 & 0.38 / 0.78&0.54 / 1.15\\
Social-STGCNN~\cite{mohamed2020social} & 0.64 / 1.11 & 0.49 / 0.85 & 0.44 / 0.79 & 0.34 / 0.53 & 0.30 / 0.48 & 0.44 / 0.75\\
STGAT~\cite{huang2019stgat} & 0.65 / 1.12 & 0.35 / 0.66  & 0.52 / 1.10 &  0.34 / 0.69 & 0.29 / 0.60 & 0.43 / 0.83\\
\hline
GraphTCN (M = 4)  & {\bf 0.59 / 1.12} & {\bf 0.27 / 0.52 }  & { 0.42 / 0.87 }  & { 0.30 / 0.62 } & { 0.23 / 0.48 } & {\bf 0.36 / 0.72 }\\
\revise{GraphTCN (M = 20)} & {\bf{0.39 / 0.71}} & {0.21 / 0.44 }  & {0.33 / 0.66 }  & {0.21 / 0.42 } & { 0.17 / 0.43 } & { 0.26 / 0.51 }\\
GraphTCN-G (M = 4)  & { 0.60 / 1.21} & {\bf 0.27 / 0.52 }  & {\bf 0.41 / 0.84 }  & {\bf 0.28 / 0.58 } & {\bf 0.22 / 0.47 } & {\bf 0.36 / 0.72 }\\
\revise{GraphTCN-G (M = 20)} & {\bf0.39} / 0.75 & {\bf 0.18 / 0.33 }  & { \bf 0.30 / 0.60 }  & { \bf 0.20 / 0.39 } & { \bf 0.16 / 0.32 } & { \bf 0.25 / 0.48 }\\
\hline
\end{tabular}
\end{center}
\vspace{-6mm}
\end{table*}

\noindent
\textbf{Ablation Study.}
We evaluate each module of GraphTCN through ablation studies in Table~\ref{ablation_study}. 
\textit{w/o EGNN} refers to the model without the spatial module.
\textit{vanilla GAT} refers to the model with GAT as the spatial module while ignoring the relative relation between pedestrians in spatial modeling.
\textit{GraphTCN-G} refers to the model which integrates VAE for multi-modal future path prediction.
The result demonstrates that introducing graph neural networks (GNN) into the framework can reduce ADE and FDE, and adding edge relative relation to GNN leads to further improvement.
However, these spatial interactions can only improve performance mildly.
We further scrutinize the dataset and attribute these findings to the fact that pedestrians seldom change their path suddenly to avoid their neighbors.
As a consequence, the temporal features already contain part of the spatial interactions for the prediction.
Therefore, spatial information is less critical in the prediction.
Meanwhile, compared with RNN-based approaches, GraphTCN can model the whole observed sequence better without losing important temporal information.

\begin{table}
\begin{center}
\caption{Ablation studies of GraphTCN.
}
\label{ablation_study}
\begin{tabular}{c||c|c}
\hline
Method & M = 4 & M = 20 \\
\hline
w/o EGNN & 0.38 / 0.78 & 0.28 / 0.54 \\
vanilla GAT & 0.37 / 0.74 & 0.27 / 0.54 \\
GraphTCN & 0.36 / 0.72 & 0.26 / 0.51 \\
GraphTCN-G & 0.36 / 0.72 & 0.25 / 0.48 \\ 
\hline

\end{tabular}
\end{center}\vspace{-3mm}
\end{table}

\noindent
\textbf{Inference Speed.}
We compare the inference speed of GraphTCN with state-of-the-art methods, including Social GAN~\cite{gupta2018social}, SR-LSTM~\cite{zhang2019sr}, Social-STGCNN~\cite{mohamed2020social} and STGAT~\cite{huang2019stgat}.
Table \ref{speed_comparison}\footnote{
For a fair comparison, the reported time includes the data processing time since some approaches require extra time to construct the graph during inference. 
Note that we use the corresponding official implementations and settings for each model, and the batch size is one in the evaluation.
} reports the model inference time and the speedup factor compared with the Social GAN in wall-clock second.
As can be observed from the results, GraphTCN achieves much faster inference compared with these baseline approaches.
In particular, GraphTCN takes 0.00067 second inference time to generate 4 samples, which is 42.82 times and 5.22 times faster than Social-GAN and the most similar prior approach STGAT respectively.

\begin{table}
\begin{center}
\caption{The inference time and speedup of GraphTCN compared with baseline methods.
The inference time is the average of the total inference steps per pedestrians.
The results are reported on an Intel Core i9-9880H Processor.}
\label{speed_comparison}
\begin{tabular}{c||c|c}
\hline
& Inference Time & Speed-up \\
\hline
Social-GAN~\cite{gupta2018social} & 0.02869 &  1$\times$\\
Social-STGCNN~\cite{mohamed2020social} & 0.00861 & 3.33 $\times$\\
STGAT~\cite{huang2019stgat} & 0.00350 & 8.20 $\times$\\
Trajectron~\cite{ivanovic2019trajectron} & 0.00081 & 35.42 $\times$\\
\hline
GraphTCN (M=4) & 0.00066 & 43.47 $\times$\\
GraphTCN-G (M=4) & 0.00067 & 42.82 $\times$\\
GraphTCN-G (M=20) & 0.00075 & 38.25 $\times$\\
\hline
\end{tabular}
\end{center}
\vspace{-3mm}
\end{table}

\subsection{Qualitative Analysis}
\begin{figure}[t]
\begin{center}
\setlength{\tabcolsep}{2pt}
\begin{tabular}{p{0.2cm}ccc}
\rotatebox[origin=lb]{90}{\hspace{1.em} STGAT} & 
\frame{\includegraphics[scale=0.1]{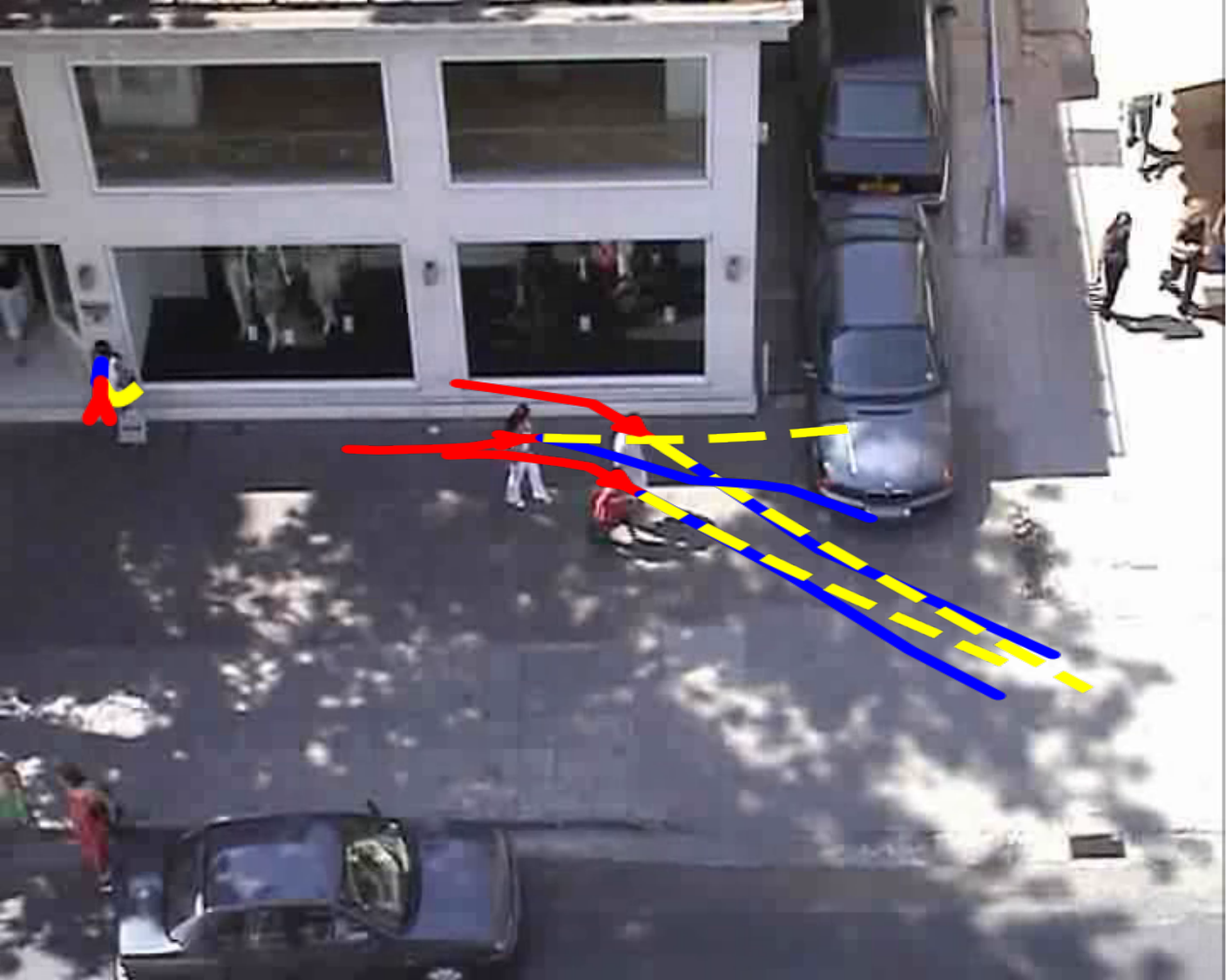}} &
\frame{\includegraphics[scale=0.1]{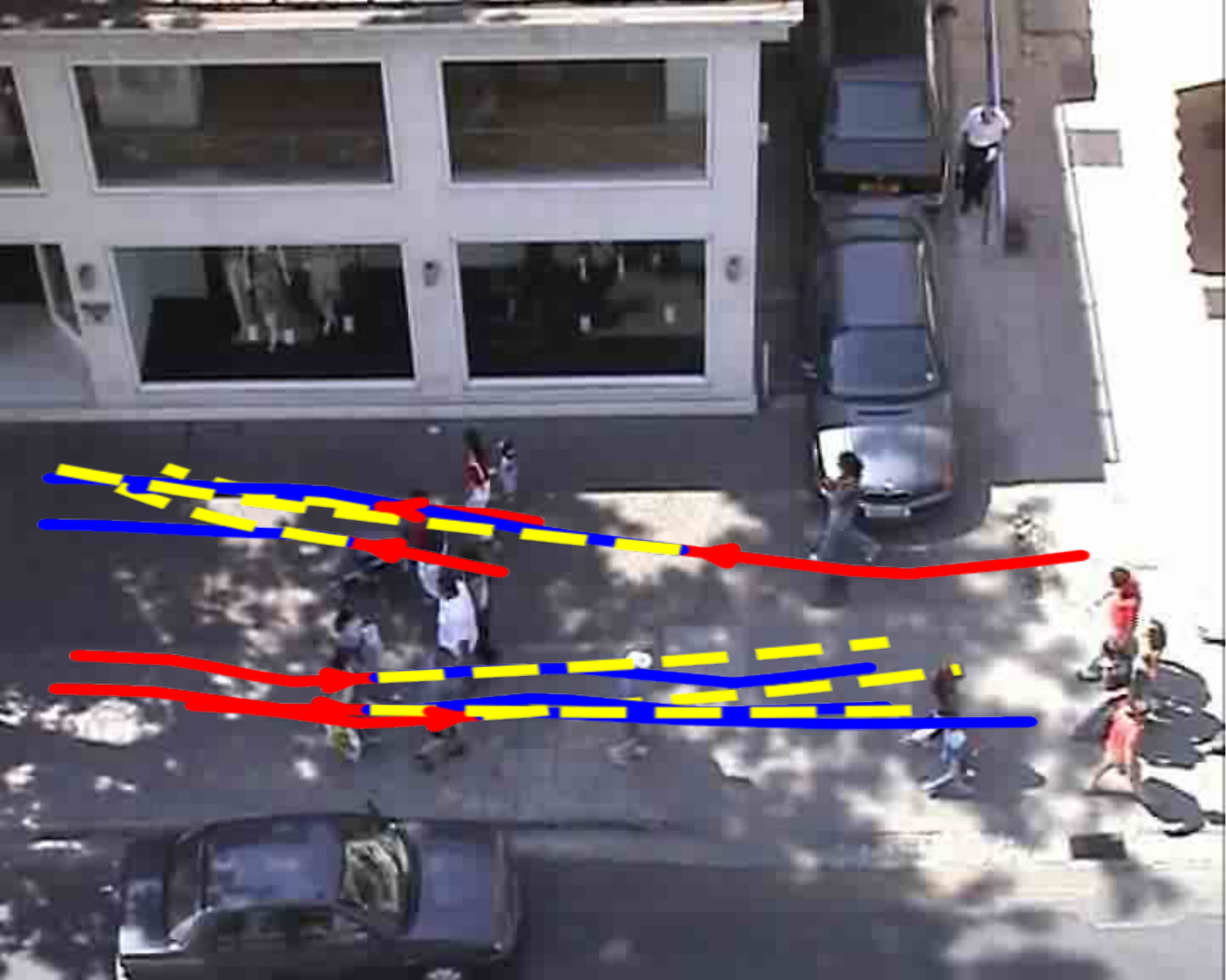}} & 
\frame{\includegraphics[scale=0.1]{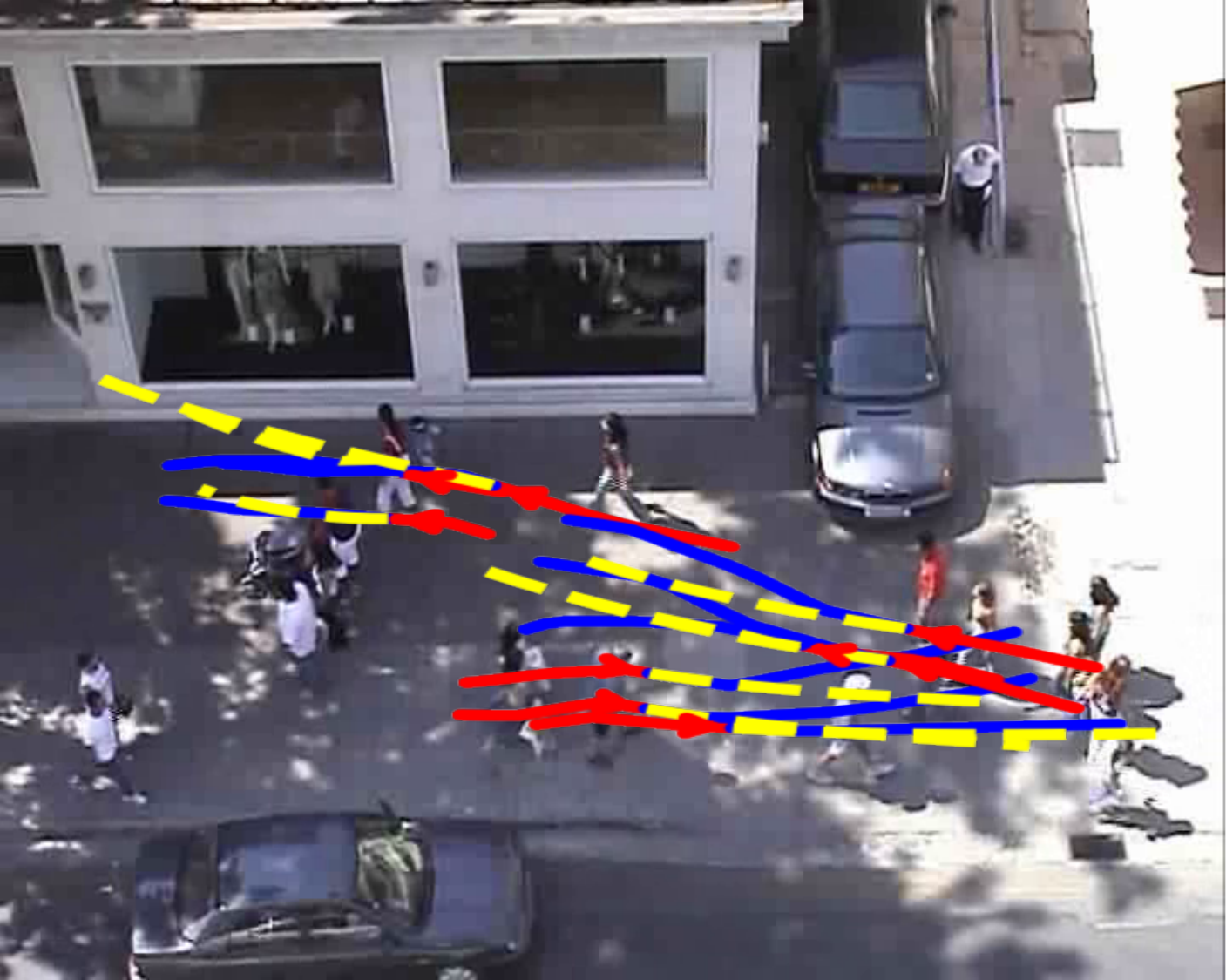}} \\
\rotatebox[origin=lb]{90}{\hspace{1.em} OURS} & 
\frame{\includegraphics[scale=0.1]{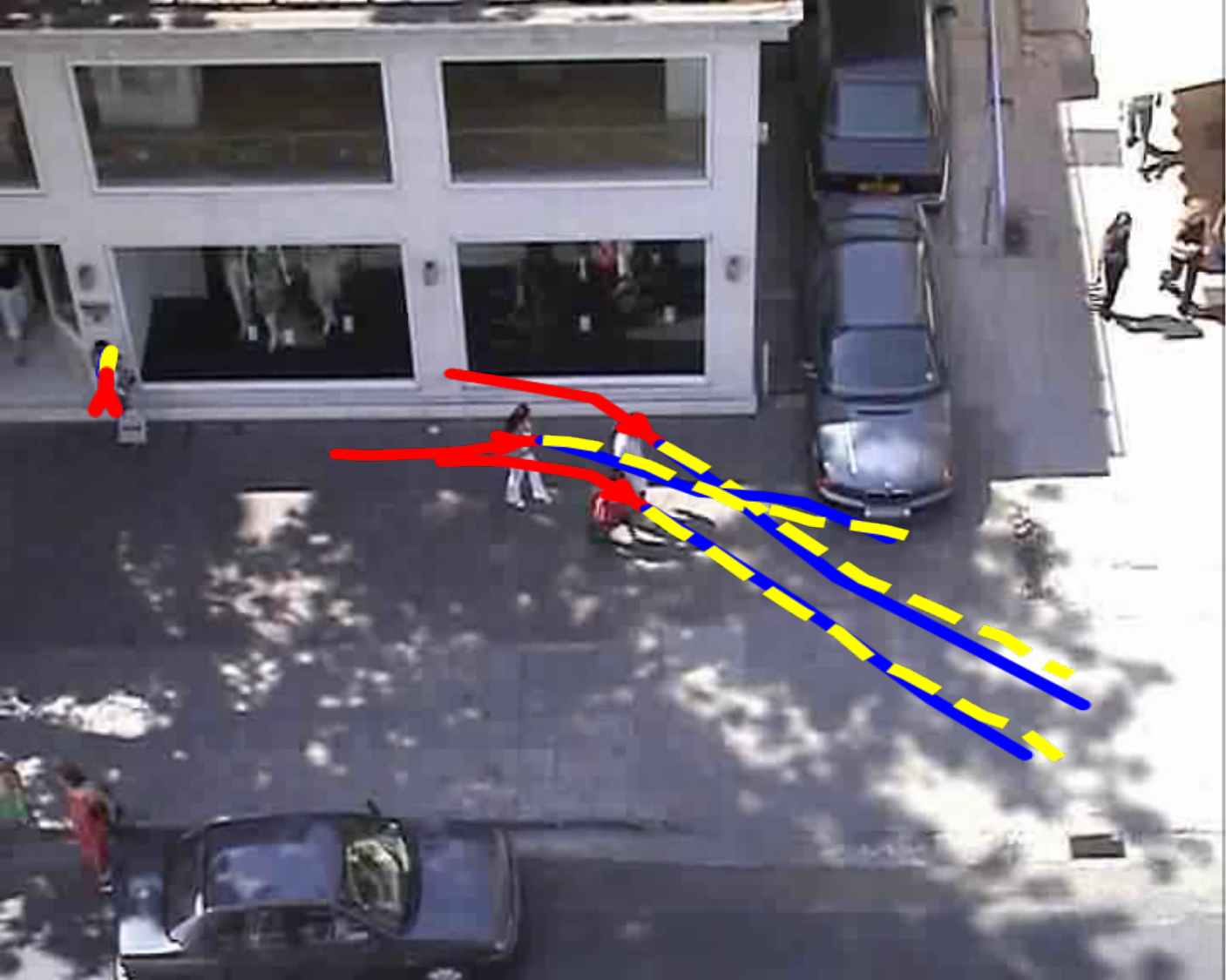}} &
\frame{\includegraphics[scale=0.1]{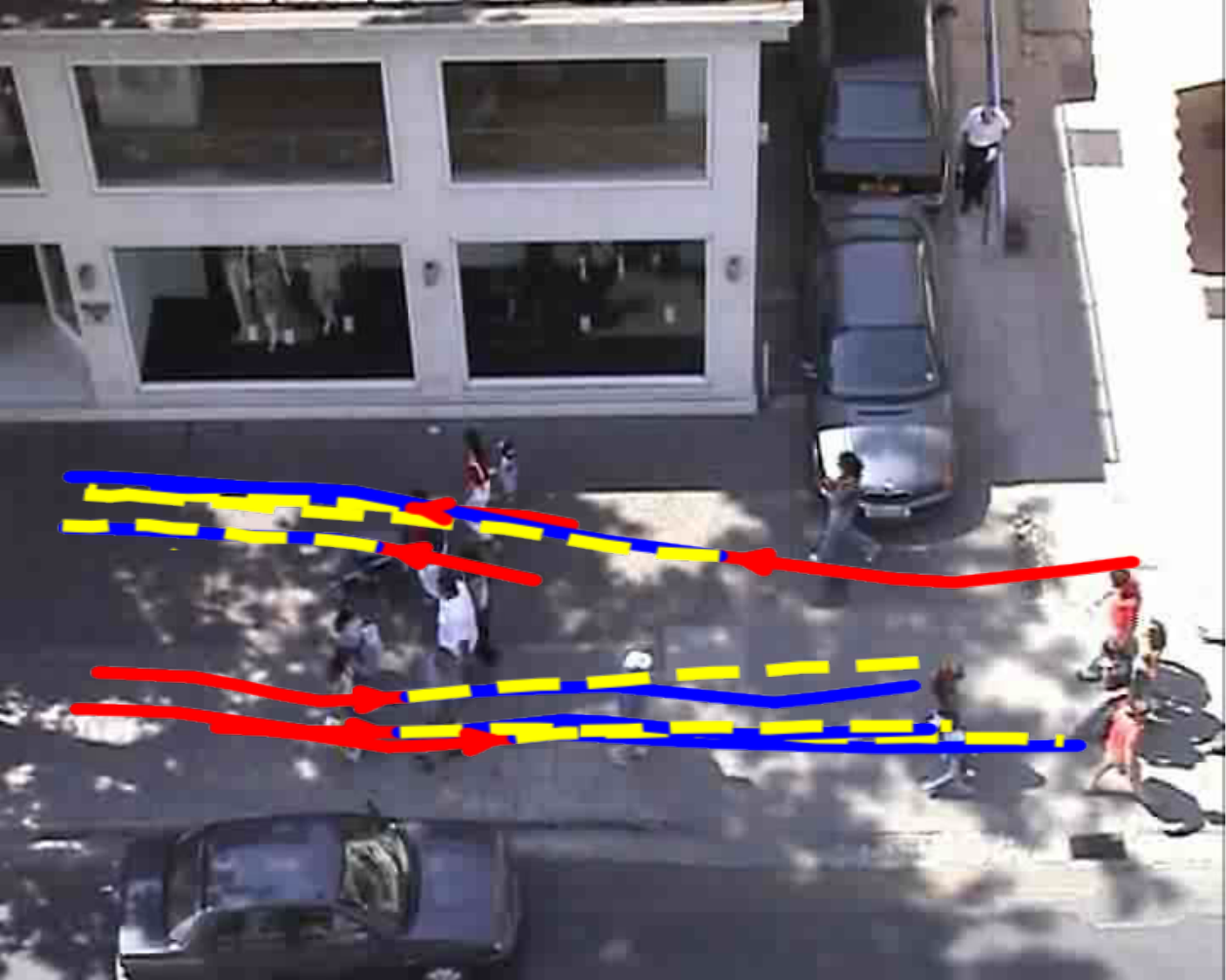}} & 
\frame{\includegraphics[scale=0.1]{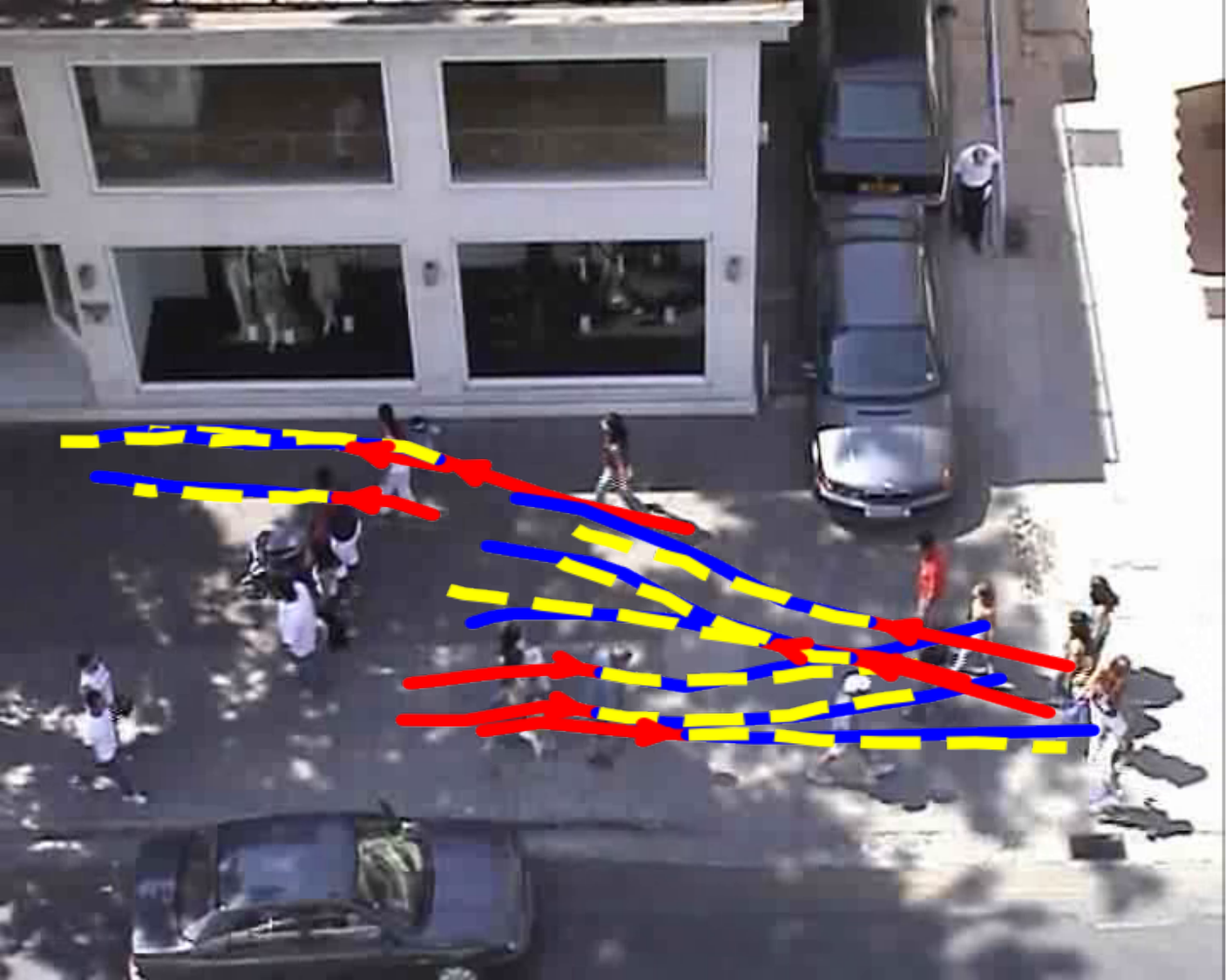}} \\
&(a)&(b)&(c)\\
\end{tabular}
\caption{Comparison of our GraphTCN (M=4) and STGAT predictions with ground truth trajectories.
To better illustrate the results, only part of the pedestrian trajectories are presented.
The solid red line, solid blue line, and dashed yellow line denote the observed trajectory, ground truth future trajectory, and predicted trajectory, respectively.}
\label{fig_quantitative_result}
\end{center}
\vspace{-0mm}
\end{figure}

We investigate the prediction results of our GraphTCN by visualizing and comparing the predicted trajectories with the best-performing approach STGAT in Fig.~\ref{fig_quantitative_result}.
We choose three different scenarios in which the complex interactions take place.
The complex interactions include pedestrian standing, pedestrian merging, pedestrian following and pedestrian avoidance. 

In Fig.~\ref{fig_quantitative_result}, we can observe that GraphTCN achieves better performance on:
firstly, \textit{direction and speed},
from Fig.~\ref{fig_quantitative_result}(a)(b), we find that trajectories generated by GraphTCN follow the same direction as the ground truth, while predictions of STGAT deviate from the path noticeably.
In Fig.~\ref{fig_quantitative_result}(a), one pedestrian moves in an unexpected direction, and GraphTCN generates an acceptable prediction accordingly.
Besides, GraphTCN generates plausible short trajectories to the stationary pedestrian and the pedestrian who moves slowly.
Secondly, \textit{collision-free future paths},
Fig.~\ref{fig_quantitative_result}(b)(c) show that STGAT may fail to make satisfactory predictions when pedestrians come from different groups, while GraphTCN generates better prediction in scenarios where one pedestrian meets another group.
In Fig.~\ref{fig_quantitative_result}(b), GraphTCN can successfully produce predictions avoiding future collisions when the pedestrian moves in the same direction from an angle.
Further, GraphTCN produces socially acceptable predictions even in the more complex scenario in Fig.~\ref{fig_quantitative_result}(c) when the pedestrian departs for the opposite directions or walks towards the same direction.


\begin{figure}[!htb]
\begin{subfigure}{.3\textwidth}
  \centering
    \frame{\includegraphics[width=1.\linewidth]{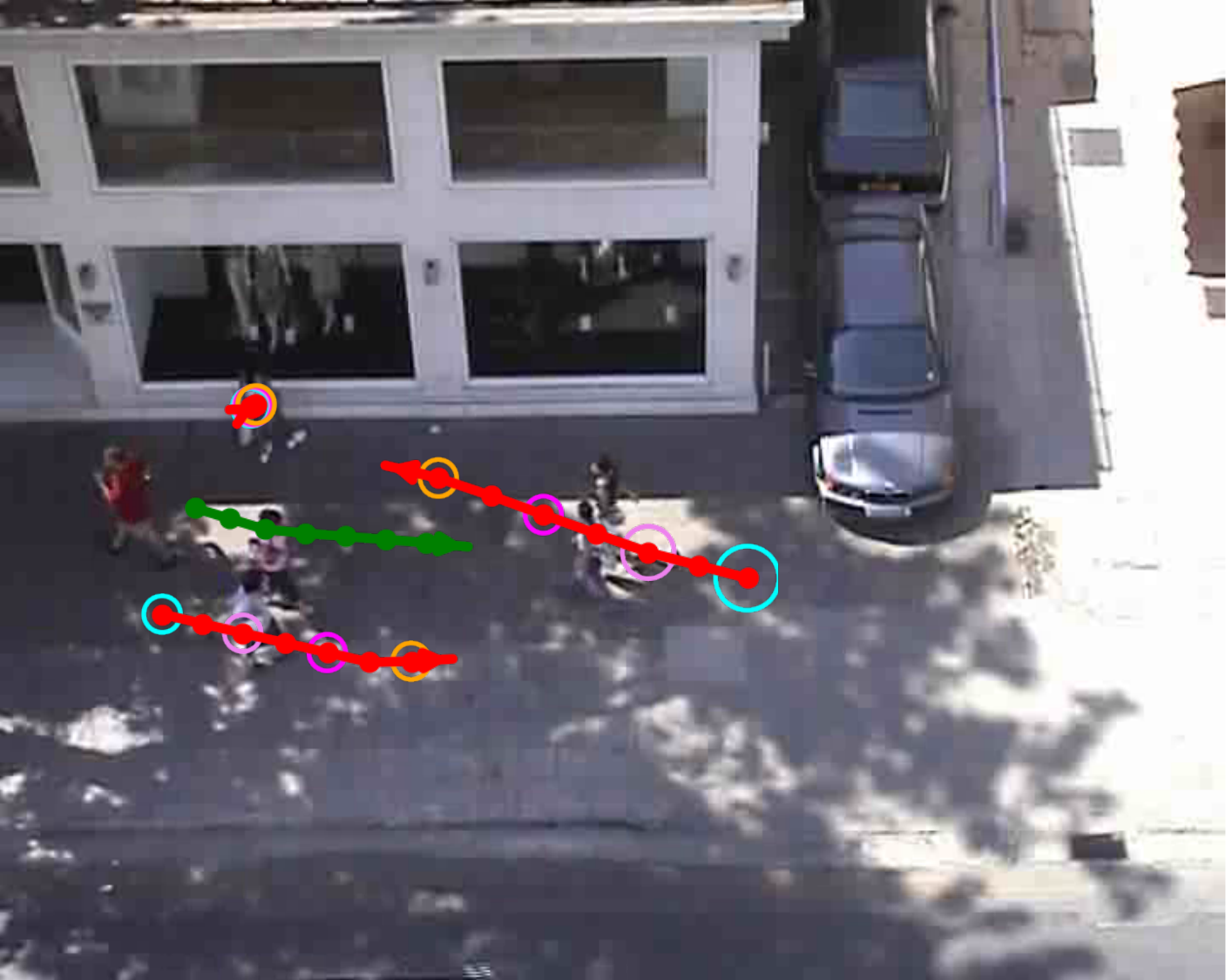}}
    \frame{\includegraphics[width=1.\linewidth]{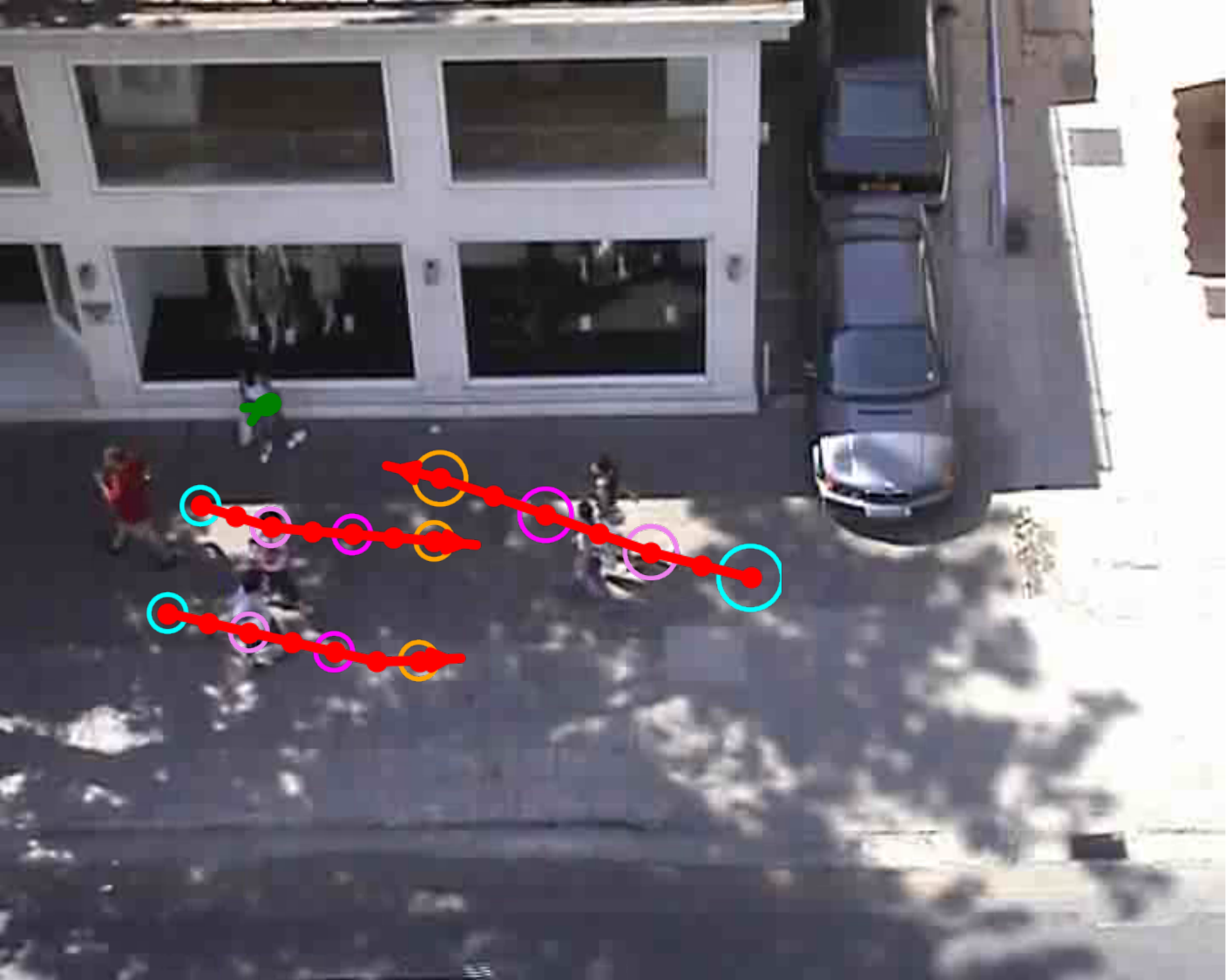}}
  \caption{}
  \label{fig:stationary}
\end{subfigure}
\begin{subfigure}{.3\textwidth}
  \centering
  \frame{\includegraphics[width=1.\linewidth]{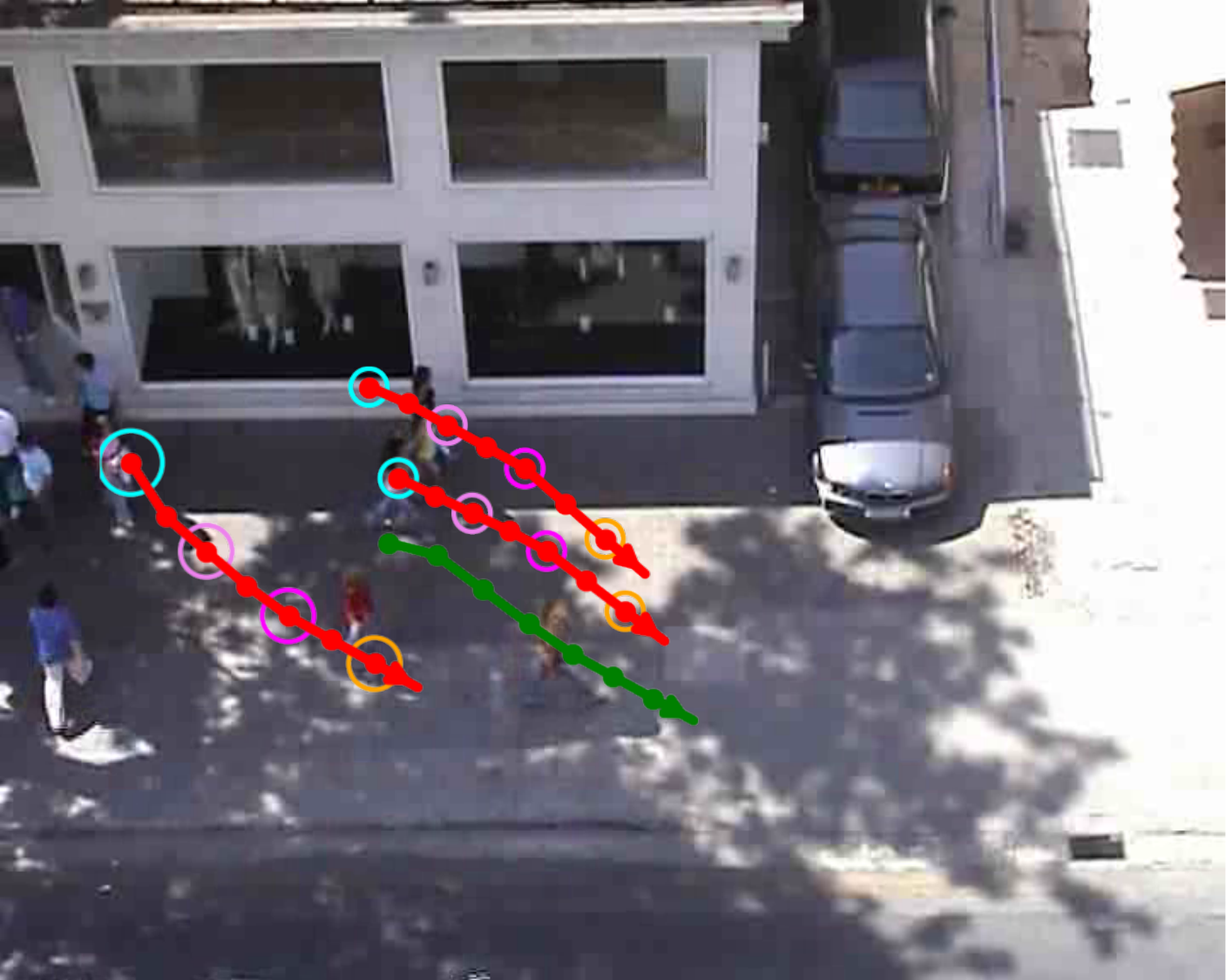}}  
  \frame{\includegraphics[width=1.\linewidth]{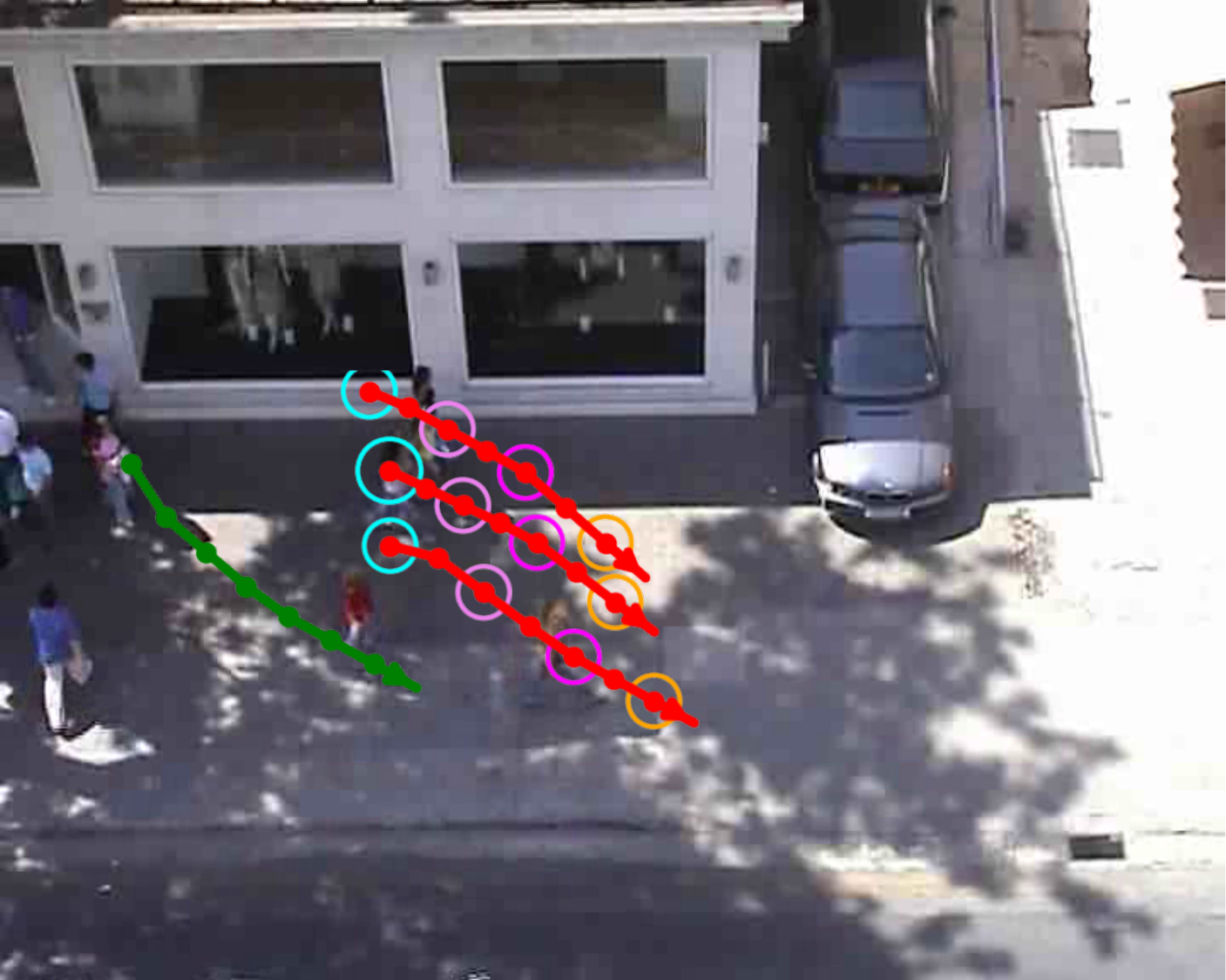}}
  \caption{}
  \label{fig:front_behind}
\end{subfigure}
\begin{subfigure}{.3\textwidth}
  \centering
  \frame{\includegraphics[width=1.\linewidth]{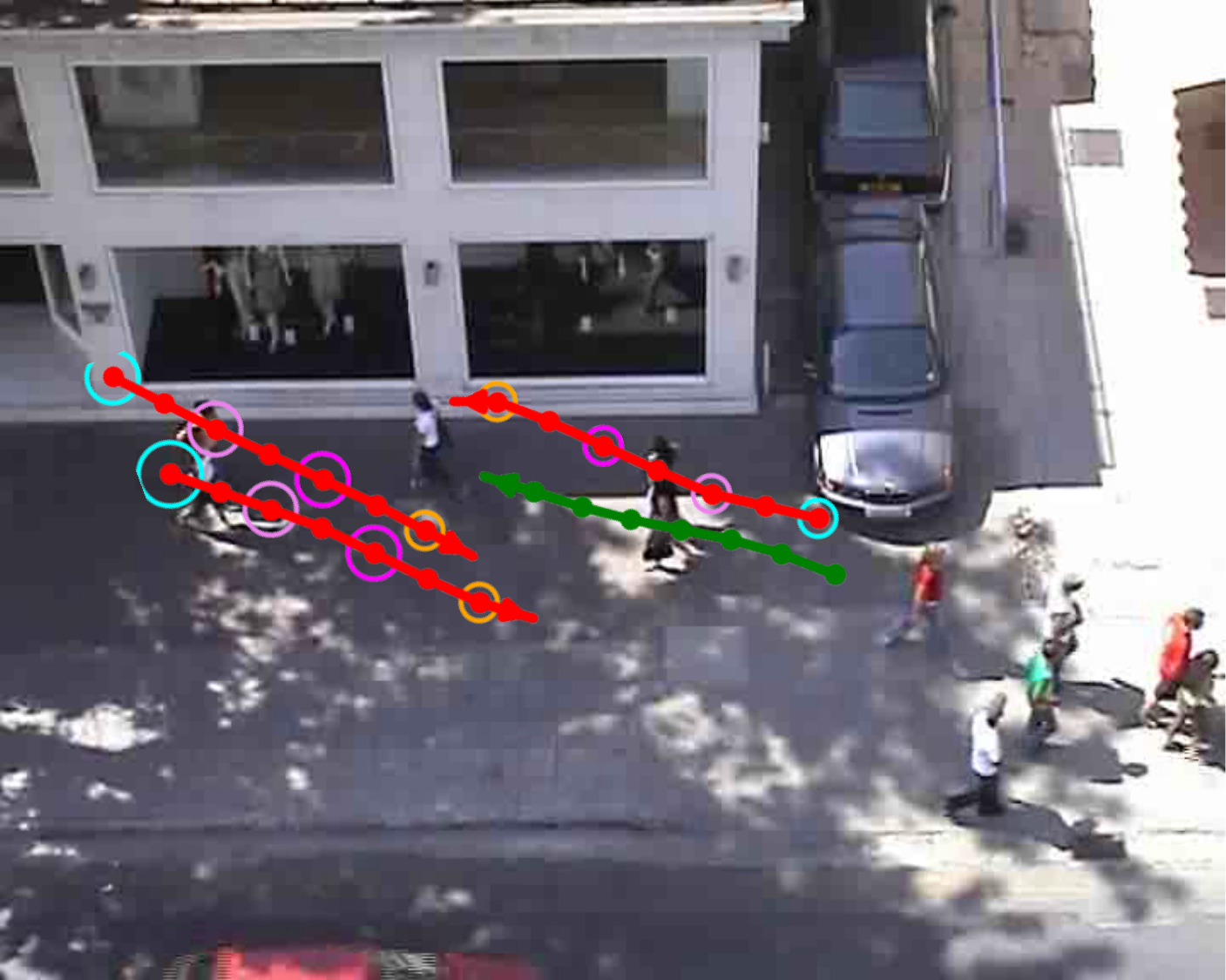}}
  \frame{\includegraphics[width=1.\linewidth]{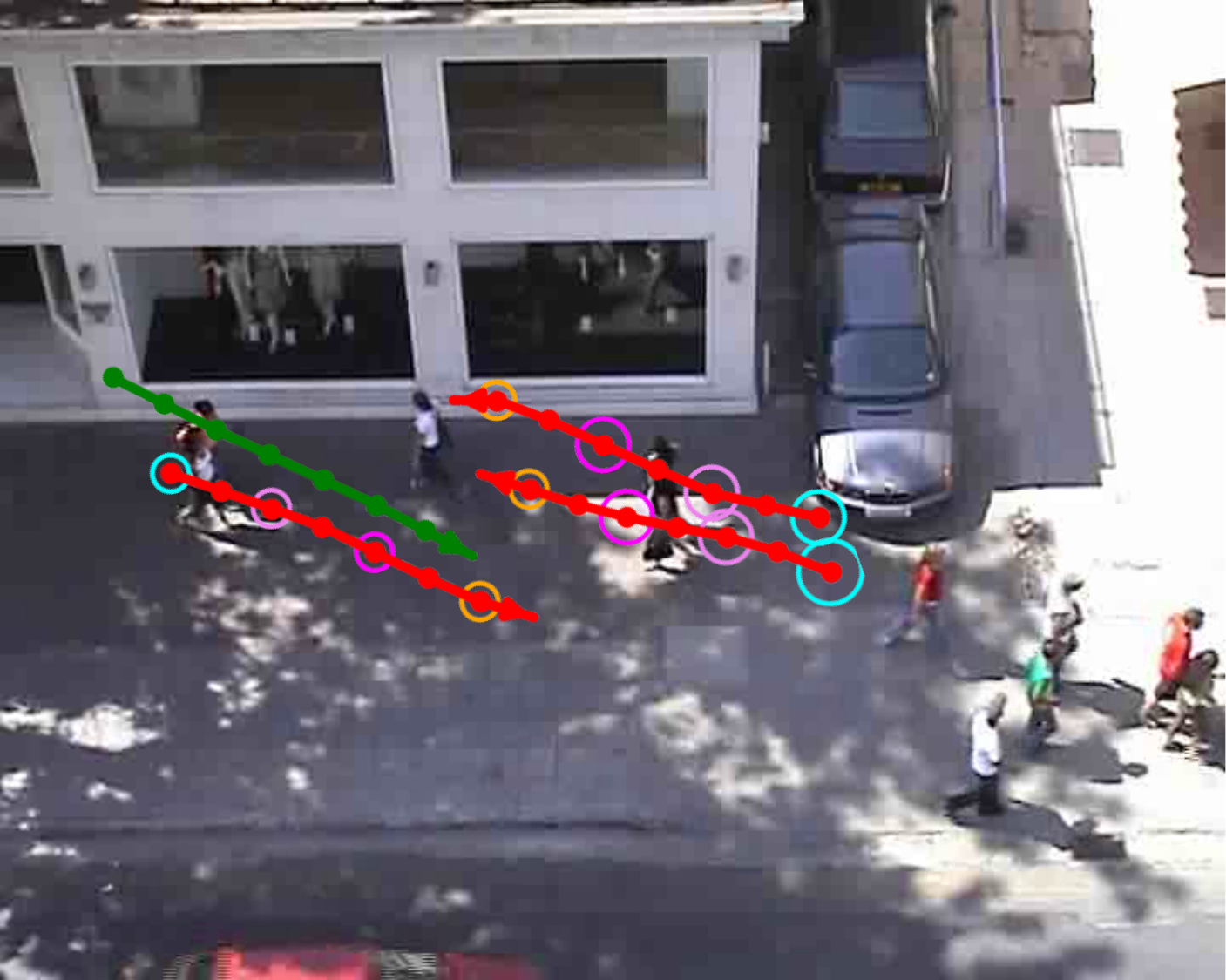}}
  \caption{}
  \label{fig:opposite_direction}
\end{subfigure}
\caption{\revise{Illustration of EFGAT attention weights.} The solid green/red line is the trajectory, and the arrow indicates the trajectory direction. 
The circle color shows the attention at each time step, and the circle size corresponds to the attention weight. 
\revise{The green trajectory without circles denotes the target pedestrian.}}
\label{fig:attn}\vspace{-0mm}
\end{figure}

\noindent
\textbf{Social attention.}
In Fig.~\ref{fig:attn}, we illustrate the learned attention weights by the EGAT module.
The results show that our model can capture the relative importance of the target's neighbors, and the attention weights between two pedestrians vary along the path.
Further, the attention weight from pedestrian $i$ to pedestrian $j$ and pedestrian $j$ to pedestrian $i$ is different, which is not considered in existing approaches~\cite{huang2019stgat,kosaraju2019social,zhang2019sr,mohamed2020social}.
\textit{Less attention weight}:
in Fig.~\ref{fig:attn}(a), the stationary pedestrian has less impact on its moving neighbors, and the model assigns small importance to the pedestrians far away from the target.
\textit{More significant influence}: our model assigns a higher attention weight to the pedestrian moving toward the target in Fig.~\ref{fig:attn}(a), moving ahead or moving in the rear while having a higher velocity in Fig.~\ref{fig:attn}(b), and moving from the opposite directions before meeting the target Fig.~\ref{fig:attn}(c).
These cases demonstrate that reasonable attention weights are successfully assigned to the target pedestrian's neighbors according to all pedestrian movement patterns in the scene.

\begin{figure}[!htb]
\begin{subfigure}{.32\textwidth}
  \centering
    \frame{\includegraphics[width=1.\linewidth]{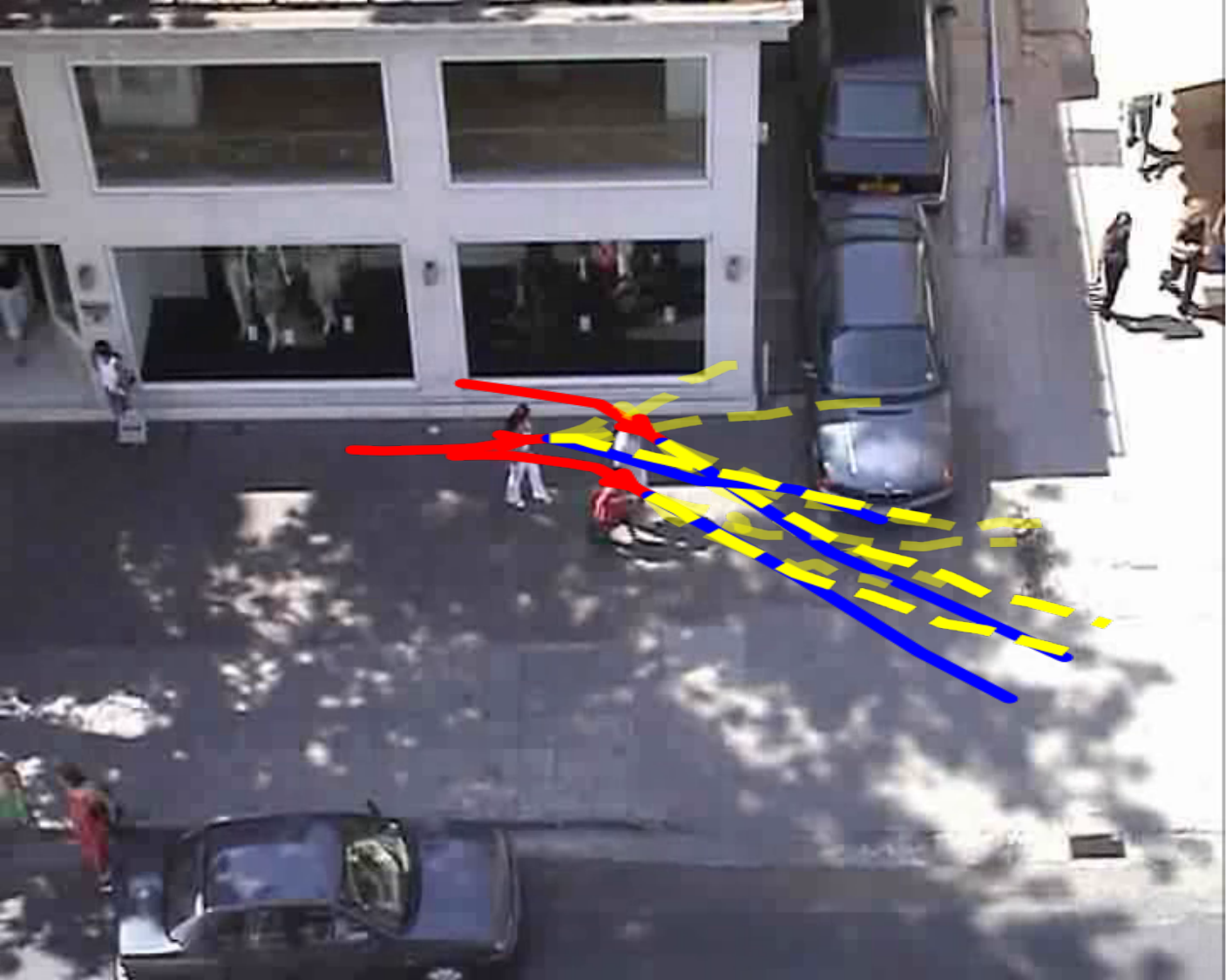}}
  \caption{GraphTCN}
  \label{fig:sample_4}
\end{subfigure}
\begin{subfigure}{.32\textwidth}
  \centering
  \frame{\includegraphics[width=1.\linewidth]{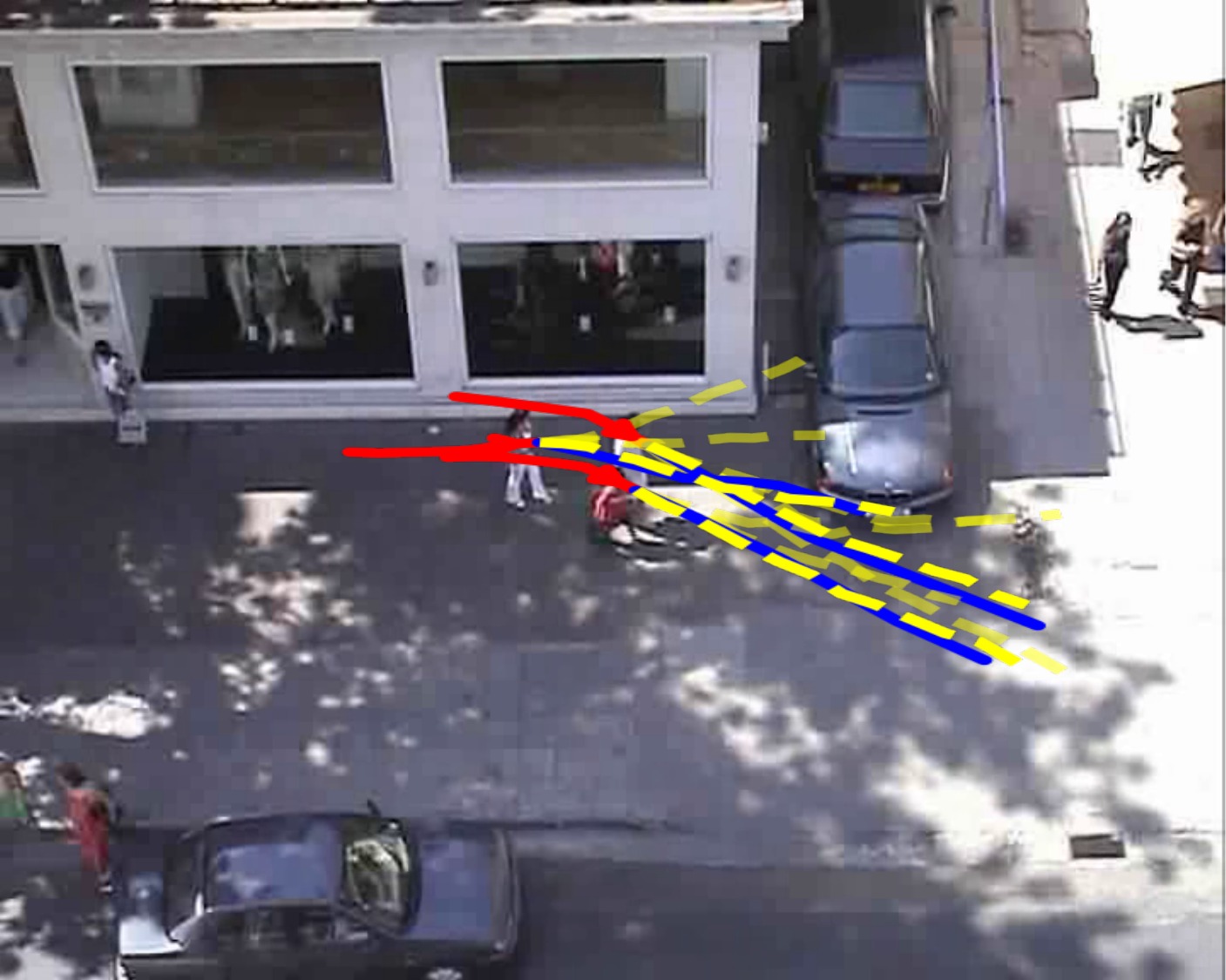}}
  \caption{GraphTCN-G}
  \label{fig:sample_20_graphtcn}
\end{subfigure}
\begin{subfigure}{.32\textwidth}
  \centering
  \frame{\includegraphics[width=1.\linewidth]{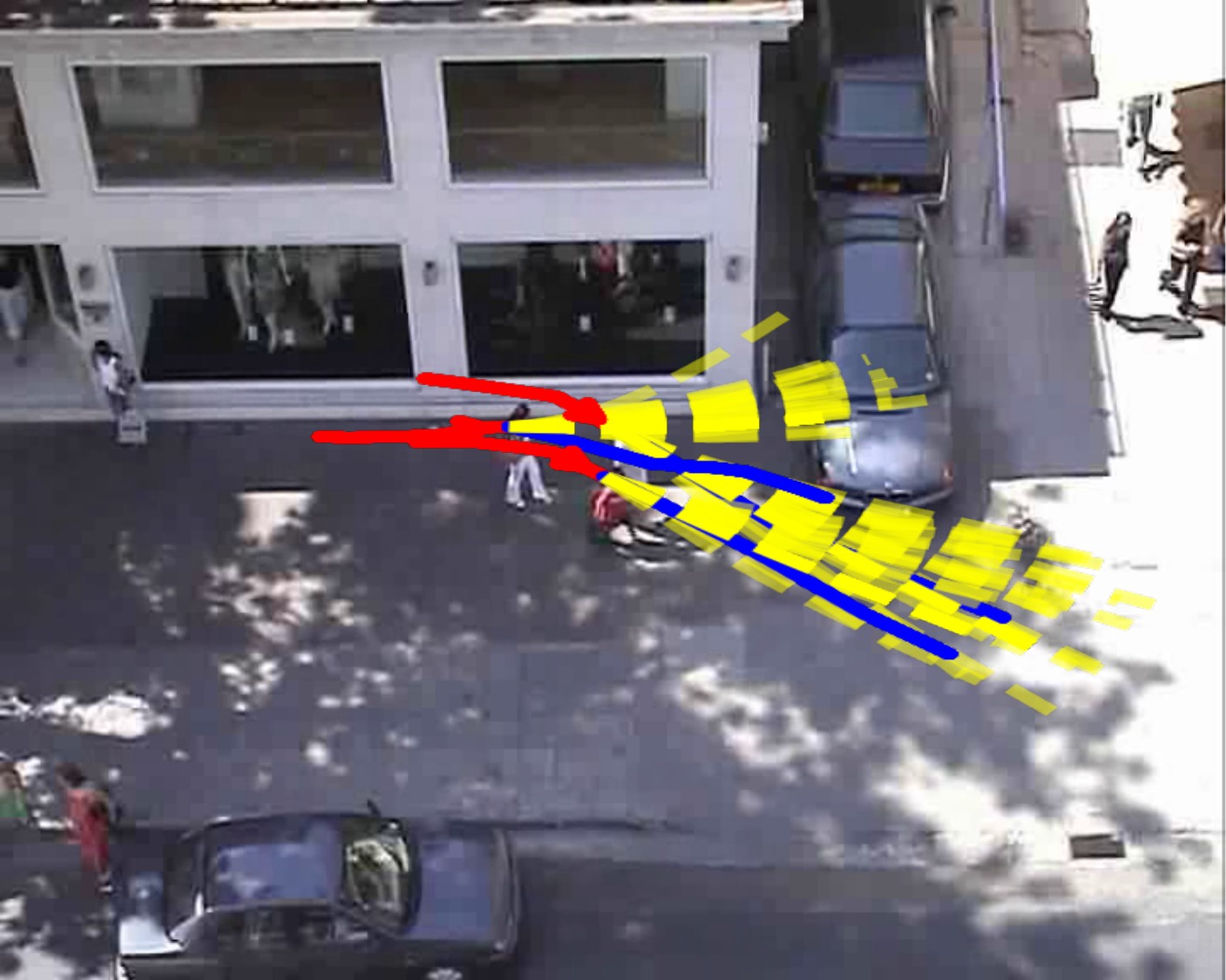}}
  \caption{STGAT}
  \label{fig:sample_20_stgat}
\end{subfigure}
\caption{Visualizations of diverse predicted trajectories. 
(a) and (b) show four trajectories produced by GraphTCN and GraphTCN-G, and (c) shows the 20 trajectories generated by STGAT.
}
\label{fig:diverse}\vspace{-0mm}
\end{figure}

\noindent
\textbf{Diverse trajectory predictions.}
Fig.~\ref{fig:diverse} is the visualization of diverse predictions.
The result shows that GraphTCN can generate the prediction closer to the ground truth even with a smaller number of samples and can make good predictions for the pedestrian, who have relatively unexpected behaviors.
In this scenario, one pedestrian has the intention to change its direction from the observation, and GraphTCN can generate both the normal and unexpected predictions for it.
And for other pedestrians who have a more consistent observation, the model produces future paths with normal behaviors.
Further, from Fig.~\ref{fig:diverse}(b) and (c), the prediction area of GraphTCN is much smaller and precise than STGAT with 20 predicted trajectories.

\section{Conclusion}
\label{sec:conclusion}

In this paper, we proposed GraphTCN for trajectory prediction, which captures the spatial and temporal interaction between pedestrians effectively by integrating EFGAT to model their spatial interactions, and TCN to model both the spatial and temporal interactions.
The proposed GraphTCN is completely based on feed-forward networks, which is more tractable during training, and achieves better prediction accuracy and higher inference speed compared with existing RNN-based solutions.
Experimental results confirm that our GraphTCN outperforms state-of-the-art approaches on various benchmark datasets.

\noindent
\textbf{Acknowledgement} This research is supported by National Research Foundation (NRF) Singapore (Award NRF2018AU-SG01).





\clearpage

{\small
\bibliographystyle{ieee_fullname}
\bibliography{egbib}
}

\end{document}